\documentclass[sigconf]{acmart}
\usepackage{multirow}
\usepackage{float}
\usepackage{amsmath}
\usepackage{geometry}
\usepackage{multirow}
\usepackage{array}
\usepackage{enumitem}
\usepackage{subcaption}
\usepackage{graphicx}
\usepackage{makecell}
\usepackage{soul, color, xcolor}
\usepackage{algorithmic} % for algorithmic environment
\usepackage{algorithm}
\usepackage{multicol}
\usepackage[bottom]{footmisc}
\newenvironment{tight_itemize}{
\begin{itemize}[leftmargin=*]
\setlength{\itemsep}{0pt}
\setlength{\parskip}{0pt}
\setlength{\parsep}{0pt}
}{\end{itemize}}

\AtBeginDocument{%
  }

\copyrightyear{2024}
\acmYear{2024}
\setcopyright{acmlicensed}\acmConference[KDD '24]{Proceedings of the 30th ACM SIGKDD Conference on Knowledge Discovery and Data Mining}{August 25--29, 2024}{Barcelona, Spain}
\acmBooktitle{Proceedings of the 30th ACM SIGKDD Conference on Knowledge Discovery and Data Mining (KDD '24), August 25--29, 2024, Barcelona, Spain}
\acmDOI{XXXXXXX.XXXXXXX}
\acmISBN{979-8-4007-0490-1/24/08}

\begin{document}

\title{Diverse Intra- and Inter-Domain Activity Style Fusion for Cross-Person Generalization in Activity Recognition}

\author{Junru Zhang}
\affiliation{%
  \institution{Zhejiang University}
  \city{Hangzhou}
  \country{China}
}
\email{junruzhang@zju.edu.cn}

\author{Lang Feng}
\affiliation{%
  \institution{Zhejiang University}
  \city{Hangzhou}
  \country{China}
}
\email{langfeng@zju.edu.cn}

\author{Zhidan Liu}
\affiliation{%
  \institution{Shenzhen University}
  \city{Shenzhen}
  \country{China}
}
\email{liuzhidan@szu.edu.cn}

\author{Yuhan Wu}
\affiliation{%
  \institution{Zhejiang University}
  \city{Hangzhou}
  \country{China}
}
\email{wuyuhan@zju.edu.cn}

\author{Yang He}
\affiliation{%
  \institution{Zhejiang University}
  \city{Hangzhou}
  \country{China}
}
\email{he_yang@zju.edu.cn}

\author{Yabo Dong}
\affiliation{%
  \institution{Zhejiang University}
  \city{Hangzhou}
  \country{China}
}
\authornote{Corresponding author}
\email{dongyb@zju.edu.cn}

\author{Duanqing Xu}
\affiliation{%
  \institution{Zhejiang University}
  \city{Hangzhou}
  \country{China}
}
\email{xdq@zju.edu.cn}

\renewcommand{\shortauthors}{Junru Zhang, et al}

\begin{abstract}
  Existing domain generalization (DG) methods for cross-person generalization tasks often face challenges in capturing intra- and inter-domain style diversity,  resulting in domain gaps with the target domain. In this study, we explore a novel perspective to tackle this problem, a process conceptualized as domain padding. This proposal aims to enrich the domain diversity by synthesizing intra- and inter-domain style data while maintaining robustness to class labels.  We instantiate this concept using a conditional diffusion model and introduce a style-fused sampling strategy to enhance data generation diversity. In contrast to traditional condition-guided sampling, our style-fused sampling strategy allows for the flexible use of one or more random styles to guide data synthesis. This feature presents a notable advancement: it allows for the maximum utilization of possible permutations and combinations among existing styles to generate a broad spectrum of new style instances. Empirical evaluations on a broad range of datasets demonstrate that our generated data achieves remarkable diversity within the domain space. Both intra- and inter-domain generated data have proven to be significant and valuable, contributing to varying degrees of performance enhancements. Notably, our approach outperforms state-of-the-art DG methods in all human activity recognition tasks.

  \end{abstract}







\maketitle

\section{introduction}
Human activity recognition (HAR) is a crucial application of time series data collected from wearable devices like smartphones and smartwatches, garnering substantial attention in recent years~\cite{lara2012survey, bulling2014tutorial, xu2021limu}. 
Deep learning (DL) techniques have proven effective in time series classification (TSC) for HAR tasks \cite{zhao2019deep, wu2022timesnet, zhang2022deep}. However, a common assumption underpinning these models is that training and test data distributions are identically and independently distributed (i.i.d.) \cite{wang2022generalizing}, a condition that does not often hold up in real life due to \emph{individual differences in activity styles} influenced by factors such as age and gender~\cite{qian2021latent, qin2023generalizable}.
For instance, sensor data distributions can diverge significantly between younger and older individuals due to variations in walking speed and frequency, leading to challenges in achieving cross-person generalization with standard DL models.

\begin{figure}[t]
  \centering
  \includegraphics[width=1\columnwidth]{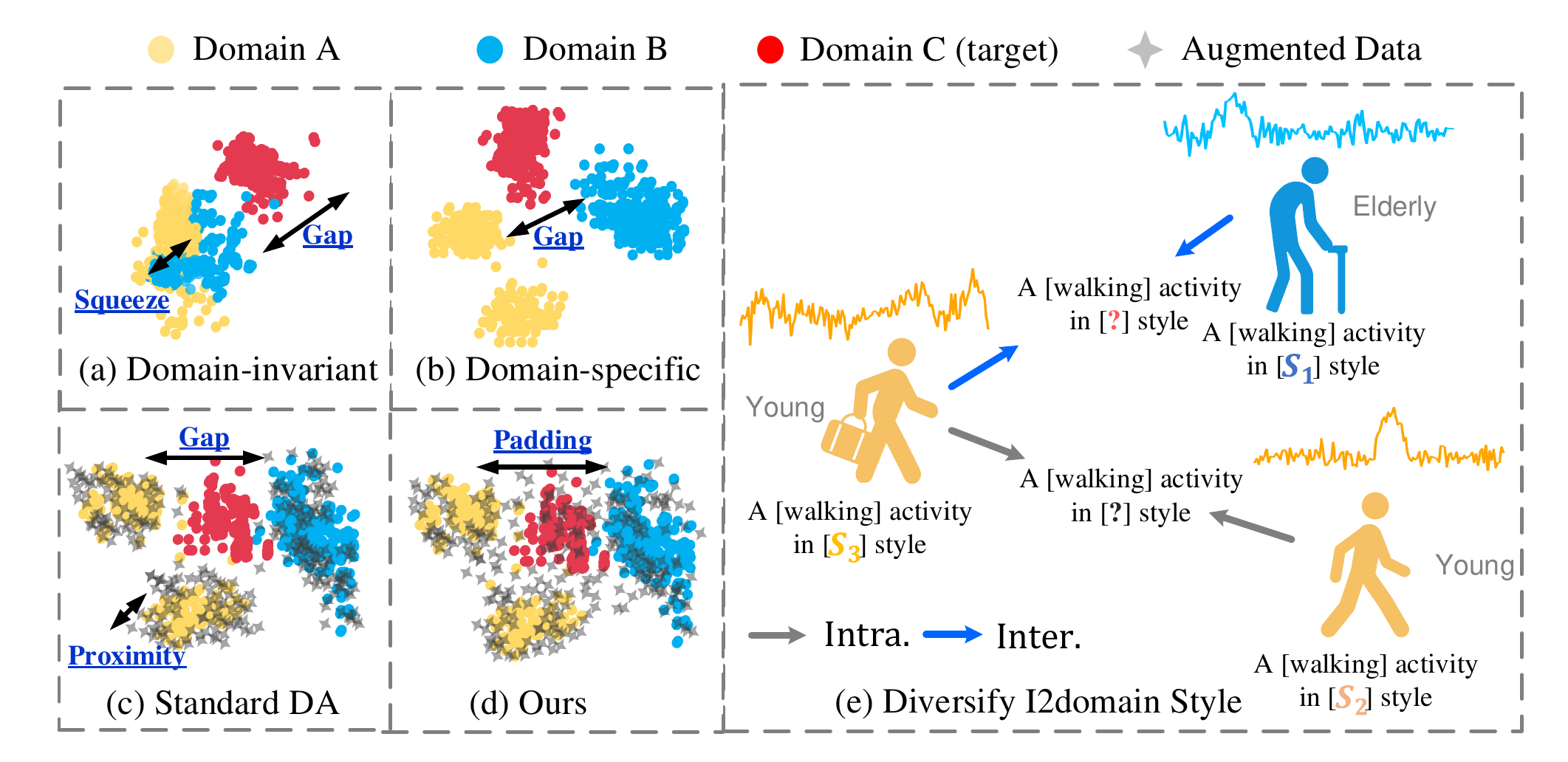}
  % \vspace{-2.5em}
  \caption{T-SNE visualization of time-series features extracted by various methods across three domains in HAR. Existing representation learning methods result in domain gaps as in both (a) and (b), covering a small portion of target domain (red circles). Standard data augmentation (DA) leads to augmented data (stars), with source domains (orange/blue circles) remaining in close proximity to each other and failing to fill gaps. Our method (d) creates a comprehensive feature space by padding domain gaps via the idea of (e).}  \label{fig:domain}
  % \vspace{-2em}
\end{figure}

Domain generalization (DG) seeks to address this issue~\cite{wang2022generalizing}. Approaches such as domain-invariant~\cite{muandet2013domain, erfani2016robust, gong2019dlow, zhou2020domain, ajakan2014domain, qian2021latent} and domain-specific~\cite{mancini2018best, wang2020dofe, zhang2023domain, bui2021exploiting} methods are designed to extract robust inter-domain and intra-domain features that can withstand data distribution shifts across various domains. However, their effectiveness is reliant on the diversity and breadth of the training data~\cite{xu2020neural}. The challenge arises in HAR tasks, where the collected training data is often small-scale and lacks the necessary diversity due to resource constraints on edge devices~\cite{qin2023generalizable, wang2023sensor}. This inherent \emph{diversity scarcity in source domain training data} can lead to overfitting to local and narrow inter- or intra-domain features, resulting in poor generalization to new, unseen domains. As shown in Fig.~\ref{fig:domain} (a) and (b), the learned features lack required intra- or inter-domain feature robustness, thereby impeding their generalization to target domains (red circles).
One promising solution is to enrich training distributions by data generation. Recent research~\cite{qin2023generalizable} has focused on enhancing training data richness through standard data augmentation like rotation and scaling; however, it primarily enhances \emph{intra-domain diversity and falls short of addressing inter-domain variability}. As shown in Fig.~\ref{fig:domain} (c), the augmented data (stars) for source domains (orange and blue circles) tends to cluster tightly, yet fails to generate the necessary inter-domain data. The target domain (red circles) thus cannot be comprehensively represented.

In this work, we focus on the generation of highly diverse data distributions to address the issue of limited domain diversity in HAR. We explore a novel perspective to tackle this problem. As depicted in Fig. \ref{fig:domain} (d), the core idea involves enabling synthetic data (stars) to fill the empty spaces within and across source domains while maintaining robustness to class labels, a process we conceptualize as ``domain padding''. For instance, as illustrated in Fig. \ref{fig:domain}(e), we can combine multiple walking styles of an elderly man and a young man to create a novel inter-domain style or merge multiple walking styles of a young man to generate a new intra-domain style. Compared to existing DG methods, our domain padding holds great potential to generate a more extensive range of unknown style distributions. This enables TSC models to comprehensively explore a wide array of intra- and inter-domain variations, contributing to enhanced generalization in HAR scenarios.

We instantiate our concept using conditional diffusion probabilistic models~\cite{sohl2015deep,ho2020denoising}. To generate samples with instance-level diversity, we first design a contrastive learning pipeline~\cite{eldele2021time}. It aims to extract the activity style representations of the available data in the source domains while preserving their robustness for classification tasks. The resulting style representation, denoted as $S_i$, can be interpreted as ``\emph{a [class] activity performed in [$S_i$] style}''. 
We then propose a novel style-fused sampling strategy for the diffusion model to achieve domain padding requirements. This involves randomly combining one or multiple style representations of training samples within the same class. Styles in each combination are then utilized to jointly guide the diffusion to generate novel activity samples that fuse the styles. This innovation presents a notable advancement: the randomness of the combination (whether originating from different or the same domains) ensures diversity in both inter-domain and intra-domain, thereby achieving the domain padding, as shown in Fig.~\ref{fig:domain} (d) and (e). Moreover, it allows for the maximum utilization of possible permutations and combinations among existing styles to generate a broad spectrum of new style instances. Hence, we term our approach as \textbf{D}iversified \textbf{I}ntra- and \textbf{I}nter-domain distributions via activity \textbf{S}tyle-fused \textbf{Diff}usion modeling (DI2SDiff). We summarize our main contributions as follows:

\begin{tight_itemize}
  \item We explore a pivotal challenge hampering the effectiveness of current DG methods in HAR: diversity scarcity of source domain features. In response, we introduce the concept of ``domain padding'', offering a fresh perspective for enhancing domain diversity and ultimately improving DG models'  performance.
  \item We propose to use activity style features as conditions to guide the diffusion process, extending the information available at the instance-level beyond mere class labels.
  \item We propose a novel style-fused sampling strategy, which can flexibly fuse one or more style conditions to generate new, unseen samples.
  This strategy achieves data synthesis diversity both within and across domains, enabling DI2SDiff to instantiate the concept of domain padding.
  \item We conduct extensive empirical evaluations of DI2SDiff across a board of HAR tasks. Our findings reveal that it markedly diversifies the intra- and inter-domain distribution without introducing class label noise. Leveraging these high-quality samples, DI2SDiff outperforms existing solutions, achieving state-of-the-art results across all cross-person activity recognition tasks.
\end{tight_itemize}

\section{Related Work}

\textbf{Human activity recognition (HAR)} uses wearable sensors for recognizing activities in healthcare and human-computer interaction \cite{chen2021deep,xu2023practically}. The complexity of daily activities, varying among individuals with different personal styles, makes recognition challenging \cite{qian2021latent,zhang2022deep}.  With the rise of deep learning, deep neural networks have been increasingly utilized to extract informative features from activity signals \cite{yang2015deep, jiang2015human, zhang2022deep, zhang2023temporal}. For example, DeepConvLSTM incorporates convolutional and LSTM units for multimodal wearable sensors \cite{qiao2020deep}. MultitaskLSTM extracts features using shared weights, then classifies activities and estimates intensity separately \cite{barut2020multitask}.

%\vspace{0.2cm}

\textbf{Domain generalization (DG)} aims to improve model performance across different domains. Early works \cite{muandet2013domain, erfani2016robust, gong2019dlow, zhou2020domain, ajakan2014domain} focused on utilizing multiple source domains and enforcing domain alignment constraints to extract robust features. For example, DANN ~\citep{ajakan2014domain} employed adversarial training to accomplish this task, but it requires target data during training.
To recall more beneficial features, several methods ~\citep{mancini2018best, wang2020dofe, zhang2023domain, bui2021exploiting} such as mDSDI ~\citep{bui2021exploiting} have been proposed to preserve domain-specific features.
Another line of research in DG focuses on data augmentation techniques ~\citep{goodfellow2014explaining, xu2020robust, volpi2018generalizing, li2021progressive, guo2023single, zheng2024advst} to explore more robust patterns for improved generalization, such as generating adversarial examples \cite{goodfellow2014explaining}. 

Given the practical significance of DG learning for HAR tasks, researchers ~\citep{wilson2020multi, qian2021latent,qin2022domain,lu2022out,qin2023generalizable} have turned their focus to studying DG problems in this field. For instance, \citet{wilson2020multi} proposed an adversarial approach to learn domain-invariant features, which requires labeled data in the target domain during training. \citet{qian2021latent} improved variational autoencoder (VAE) framework \cite{kingma2013auto} to disentangle domain-agnostic and domain-specific features automatically, but domain labels are required.  DDLearn ~\citep{qin2023generalizable} is a recent advanced approach that enriches feature diversity by contrasting augmented views but is limited to standard augmentation techniques that only enrich intra-domain features.

%\vspace{0.2cm}
\textbf{Diffusion models} have showcased their remarkable potential in generating diverse and high-quality samples in various domains, like computer vision~\cite{lugmayr2022repaint}, natural language processing~\cite{li2022diffusion}, and decision-making~\cite{feng2024resisting}. Furthermore, classifier-free guidance models \cite{ho2022classifier} have achieved impressive outcomes in multimodal modeling, with wide applications in tasks such as text-to-image synthesis \cite{rombach2022high} and text-to-motion \cite{tevet2022human}.
Considering the potential non-stationary distribution of time-series data \cite{huang2020causal}, we propose harnessing the power of diffusion models to generate diverse data in HAR tasks, and thereby enhancing the model's generalization ability. Intriguingly, diffusion models have received limited attention in HAR tasks. A recent survey on time-series diffusion models \cite{lin2023diffusion} indicates that although some successful attempts have been made to apply diffusion models to time-series tasks like interpolation \cite{tashiro2021csdi} and forecasting \cite{rasul2021autoregressive,bilovs2022modeling}, comprehensive investigations in time-series generation tasks are still lacking. %We address this gap by not only establishing diffusion models for time-series generation but also guiding the diffusion model to generate diverse samples to address DG issues in HAR tasks. This makes our work  novel and challenging.
Our study not only establishes diffusion models for time-series generation but also guides the diffusion model to produce diverse samples, effectively addressing the challenges of DG in HAR tasks. Our work thus presents a novel and challenging contribution to the field.

%\vspace{-1.5em}

\section{Preliminaries}

\subsection{Problem Statement}
In cross-person activity recognition~\cite{qin2023generalizable}, a domain is characterized by a joint probability distribution $P_{X,Y}$ across the product space of time-series instances $\mathcal{X}$, and the corresponding label space $\mathcal{Y}$.
Each instance $\mathbf{X}_i \in \mathbb{R}^{K \times L}$ represents the values of each time series obtained from sensors, where $K$ is the dimensionality of features, and $L$ is the temporal length of the series. Moreover, each instance $\mathbf{X}_i$ corresponds to an activity class label $y_i \in \{1, 2, \dots, C\}$, indicating the specific activity category performed by the subjects, with $C$ denoting the total number of activity categories.
The domain generalization challenge lies in the nonintersection and domain differences between the training and testing sets. Typically, the training set $D^s=\{(\mathbf{X}_i, y_i)\}_{i=1}^{n^s}$ is collected from the labeled source-domain subjects, where $n^s$ represents the number of training instances. Importantly, $n^s$ is often small in cross-person activity recognition scenarios, presenting the \emph{small-scale} challenge. On the other hand, the test set $D^t=\{(\mathbf{X}_i, y_i)\}_{i=1}^{n^t}$ consists of $n^t$ instances obtained from the unseen target-domain subjects and satisfies the condition $D^s \cap D^t = \emptyset$. In addition, source and target domains have different joint probability distributions while sharing the identical feature space and class label space, i.e., $P^s(\mathbf{X}_i, y_i) \neq P^t(\mathbf{X}_i, y_i)$, and $ \mathcal{X}^s =  \mathcal{X}^t$, $ \mathcal{Y}^s =  \mathcal{Y}^t$.
The primary objective is to leverage the available data in $D^s$ to train a TSC model $f : \mathcal{X} \rightarrow \mathcal{Y}$ capable of effectively generalizing to an inaccessible, unseen test domain $D^t$, without any prior exposure to target domain data or domain labels during training. This task is inherently more challenging than conventional transfer learning settings~\cite{qin2023generalizable, qian2021latent} due to the disparate distributions across source and target domains, compounded by the small-scale settings of the training data.

%\vspace{-1em}

\subsection{Diffusion Probabilistic Model} \label{background}

Diffusion model~\cite{sohl2015deep} involves training a model distribution $p_{\theta}(x)$ to closely approximate the target ground-truth data distribution $q(x)$. It assumes distribution $p_{\theta}(x)$ as a Markov chain of Gaussian transitions: $p_{\theta}(x_0)=\int p_{\theta}(x_T)\prod_{t=1}^{T} p_{\theta}(x_{t-1} | x_t)dx_{1:K}$, where $x_1,\dots,x_T$ denote the latent variables with the same dimensionality as original (noiseless) data $x_0$. $p_{\theta}(x_T)\sim \mathcal{N}(0, \textbf{I})$ is the Gaussian prior. $p_{\theta}(x_{t-1} | x_t)$ is the trainable reverse process given by
\begin{equation}
  p_{\theta}(x_{t-1} | x_t) := \mathcal{N}(x_{t-1}; \mu_{\theta}(x_t, t), \sigma_{\theta}(x_t, t)).
  \label{reverse1}
\end{equation}
Diffusion predefines a forward process that progressively adds Gaussian noise to $x_0 $ in $T$ steps, defined as
\begin{equation}
  \label{eq:forward}
  q(x_t | x_{t-1}) := \mathcal{N}\left(x_{t}; \sqrt{1-\beta_t} x_{t-1}, \beta_t \textbf{I}\right),
\end{equation}
where $\beta_t\in(0,1)$ is variance schedule for noise control.

\textbf{Training procedure.}~~
The diffusion model's parameters $\theta$ are optimized by maximizing the evidence lower bound of the log-likelihood of the data, i.e., $\log p_{\theta}(x_0)$, which can be further simplified as a surrogate loss~\cite{ho2020denoising}:
\begin{align}
  \mathcal{L}(\theta) := \mathbb{E}_{{x}_0,t\sim \mathcal{U},{\epsilon} \sim \mathcal{N}(0, \textbf{I}) } \left[|| {\epsilon} - {\epsilon}_{\theta}(x_t, t) ||^2\right],
  \label{eq:loss}
\end{align}
where $\mathcal{U}$ is the uniform distribution and the noise predictor ${\epsilon}_{\theta}(x_t, t)$, parameterized with a deep neural network, aims to estimate the noise $\epsilon$ at time $t$ given $x_t$. As $\mu_{\theta}(x_t, t)$ is determined by ${\epsilon}_{\theta}(x_t, t)$, the target $p_{\theta}(x_{t-1} | x_t)$ can be consequently  derived.

\textbf{Sampling procedure.}~~
Given a well-trained $p_{\theta}$, the data generation procedure begins with a Gaussian noise $x_T \sim \mathcal{N}(0, \textbf{I})$ and proceeds by iteratively denoising $x_{t}$ for $t=T,\dots,1$ through $p_{\theta}(x_{t-1}|x_t)$, culminating in the generation of the new data $x_0$.

\section{Domain Padding} \label{condition}

A major obstacle to achieving domain generalization in HAR tasks is the limited data diversity of the source domain. This presents representation learning methods from extracting robust features to distribution shifts across domains. Moreover, data augmentation also shows insufficient data richness within the domain space, particularly the inter-domain distribution.

In response, our work aims to achieve highly diverse data generation to enrich the training distributions. We propose a novel perspective, which we refer to as ``domain padding''.
The core idea is to achieve the richness of the domain space by ``padding'' the distributional gaps within and between source domains, as demonstrated in Fig.~\ref{fig:domain} (d).
To ensure the generation of high-quality, diverse data that can effectively augment the training datasets for HAR models, domain padding adheres to two key criteria:
\begin{tight_itemize}
    \item \textbf{Class-Preserved Generation}: The generated data should maintain alignment with the original data in terms of class labels, ensuring consistency.
    \item \textbf{ Intra- and Inter-Domain Diversity}: The generated data should not only boost diversity within individual domains (i.e., intra-domain diversity) but should also enrich distribution between distinct domains (i.e., inter-domain diversity). 
\end{tight_itemize}

The first criterion ensures that the enhanced diversity does not compromise the semantic integrity of the data. By maintaining consistency in class labels, domain padding contributes meaningfully to model learning without introducing label noise or confusion. The second criterion guarantees that the models are exposed to a wide range of domain variations, thereby enhancing their robustness against shifts in data distribution.

\section{Methodology}
We implement domain padding using conditional diffusion models given their highly-expressive generative capabilities \cite{ho2020denoising, yang2023diffusion}. The iterative denoising process of diffusion models makes them exceptionally suited for flexible conditioning mechanisms.
In this framework, given the original dataset ${D}^s$, we generate a new sample $\tilde{x}_0 \sim \tilde{\mathcal{X}}^s$ using conditional information $s \in \mathcal{X}^{\text{cond}}$ to guide the generation process. The ensemble of all generated data constitutes a synthetic dataset, denoted as $\tilde{D}^s= \{({\mathbf{\tilde{X}_i}}, {{y}_i})\}_{i=1}^{\tilde{n}^s}$, with $\tilde{n}^s$ denoting the total count of generated samples. Next, we use $\tilde{x}_0 $ to denote an example of the synthetic samples. The generation objective is to estimate the conditional data distribution $q(\tilde{x}|s)$.
This allows us to generate a synthetic sample $\tilde{x}_0$ given a specific constraint $s$. The conditional diffusion process can be described by:
\begin{equation}
\label{eq:cond}
q(\tilde{x}_{t} |\tilde{x}_{t-1}, s ), \quad p_\theta(\tilde{x}_{t-1} |\tilde{x}_{t}, s).
\end{equation}
Sequentially performing $p_{\theta}$ enables the generation of new samples to capture the attributes of $s$. However, realizing domain padding is not a trivial task due to a key aspect: \emph{how to guide the diffusion model to generate diverse activity samples meeting two criteria of domain padding}. 

In the following, we introduce the DI2SDiff framework, designed to enable diffusion to achieve domain padding. In \S \ref{cond_sec}, we present a contrastive learning pipeline that extracts style features to serve as conditions for the diffusion model. Given a style condition, we employ classifier-free guidance~\cite{ho2022classifier} to generate new samples that meet the first criterion in \S \ref{class_sec}. For the second criterion, we construct a diverse style combination space for the condition space $\mathcal{X}^{\text{cond}}$ and introduce a style-fused sampling strategy to generate highly diverse intra and inter-domain data in \S \ref{multi_sec}. We finally provide the workflow of DI2SDiff in \S \ref{detail_sec}.

\subsection{Activity Style Condition}\label{cond_sec}

Conditional diffusion models are typically guided by label or text prompts that provide task-specific knowledge, such as ``create a [cartoonish] [cat] image'' \cite{min2023recent, zhang2022glipv2, wang2022learning}. However, the generation of instance-level time-series data introduces distinct challenges. It is difficult to capture the complex patterns solely through label or text prompts due to the inherently high-dimensional and non-stationary nature~\cite{huang2020causal}. 
To address this issue, we propose the development of a style conditioner using a contrastive learning approach~\cite{eldele2021time}. This approach has demonstrated robustness in extracting representations from unlabeled time-series data. The transformed data can retain the distinctive characteristics of the original data while preserving the semantic information of the classes. Thus, it is well-suited for extracting robust \emph{instance-level representation}, termed as ``\emph{style}'', which can serve as conditions to guide diffusion models.

Delving into specifics, the contrastive learning pipeline consists of a feature encoder and Transformer on the available training data. The objective is to maximize the similarity between different contexts of the same sample and minimize the similarity between contexts of different samples.
Once the module is trained, we utilize it as \emph{style conditioner} denoted as $f_{\text{style}}$. When extracting the style from the original data $\mathbf{X}_i$, the style conditioner produces a style vector $S_i = f_{\text{style}}(\mathbf{X}_i) \in \mathbb{R}^H$, where $H$ denotes the length of the vector. Consequently, each activity style condition can be interpreted as ``\emph{a [$y_i$] activity performed in [${S}_i$] style}'', where $y_i$ denotes the class of the original data. This approach takes an important step towards the first criteria of domain padding due to the preservation of class semantics. The aggregation of all context vectors from $n^s$ training instances constitutes a set $\mathcal{S} =\{ {S_i} \}_{i=1}^{n^s}$, which can be further divided into $C$ class-specific subsets corresponding to $C$ classes. Each subset contains style vectors pertaining to a specific class, expressed as $\mathcal{S} =\{ \mathcal{S}^1 \cup \mathcal{S}^2 \cup \cdots \cup \mathcal{S}^C \}$. 
In Appendix \ref{append:sty}, we provide the details of the contrastive learning approach~\cite{eldele2021time}.

\begin{figure*}
  \centering
  \includegraphics[width=0.8\linewidth]{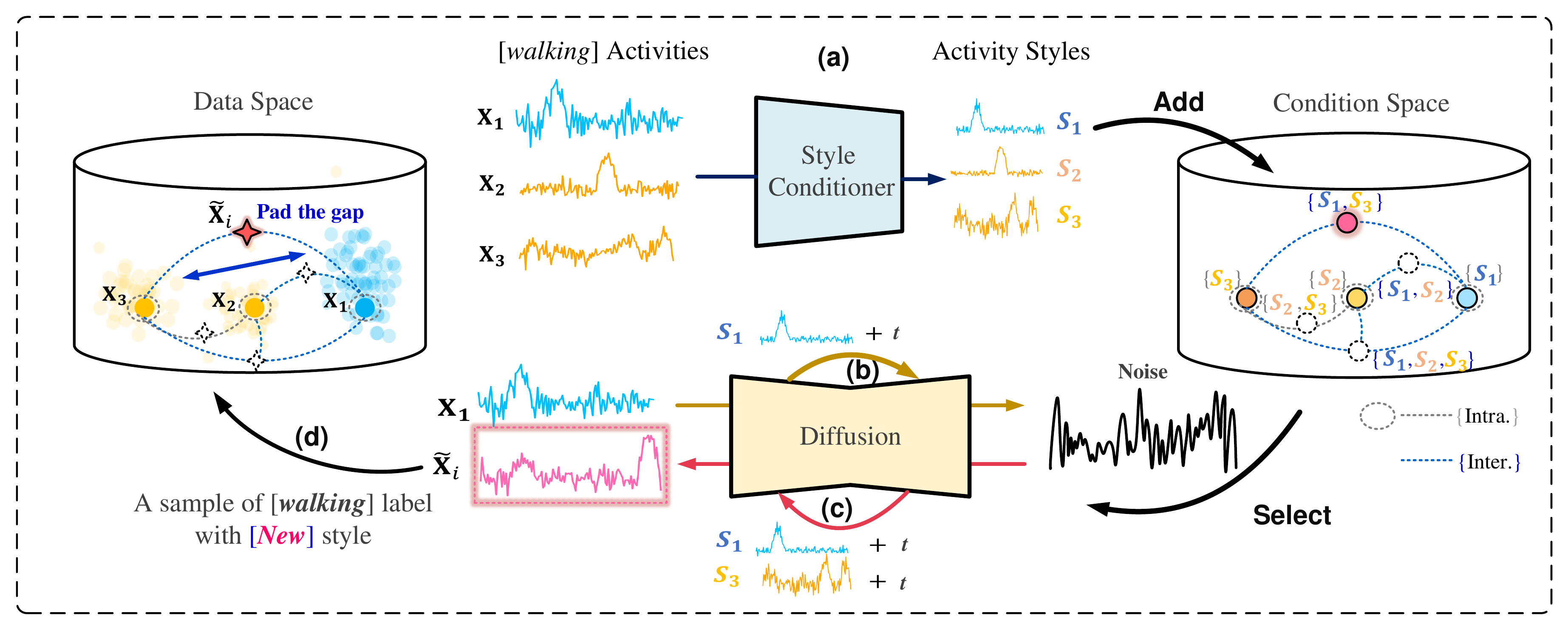}
  % \vspace{-1em}
  \caption{Illustration of the diffusion within DI2SDiff. It contains a style conditioner to produce styles and a conditional diffusion for data generation. Suppose we have three original walking samples: $\mathbf{X}_1$, $\mathbf{X}_2$, and $\mathbf{X}_3$, where $\mathbf{X}_1$ is from a different domain while $\mathbf{X}_2$ and $\mathbf{X}_3$ come from the same domain.
  (a) The style conditioner generates style features from the original data. The style features are randomly combined to build the condition space, in which the combination of inter-domain styles is indicated by blue brackets and the combination of intra-domain styles is indicated by grey brackets. 
  (b) During training, the diffusion retrieves each data sample with one style for the forward process.
  (c) During sampling, the diffusion receives noise and a style combination, e.g., $[S_1, S_3]$, for the reverse process.
  (d) The generated sample $\tilde{\mathbf{X}}_i$ is used to diversify the data space. 
  }  \label{all}
  % \vspace{-0.5em}
\end{figure*}

\subsection{Synthesizing with Classifier-Free Guidance} \label{class_sec}

To control the generation of time-series samples, we can leverage the style in $\mathcal{S}$ to guide the conditional sampling process $p_\theta(\tilde{x}_{t-1} |\tilde{x}_{t}, s)$ presented in Eq.~(\ref{eq:cond}). To this end, we adopt the classifier-free guidance~\cite{ho2022classifier}, which has proven to be effective in generating data with specific characteristics. In this framework, the training process is modified to learn a conditional $\epsilon_\theta(\tilde{x}_t, t, s)$ and an unconditional $\epsilon_\theta(\tilde{x}_t, t, \emptyset)$, where $\emptyset$ symbolizes the absence of the condition $s$. The loss function is formulated as follows:
\begin{align}
  \mathcal{L}(\theta) &:=  \mathbb{E}_{{x}_0 \sim \mathcal{X}^s, \epsilon \sim \mathcal{N}(0, \textbf{\textit{I}}), t\sim \mathcal{U}, s \sim \mathcal{S}} \left[\| \epsilon - \epsilon_\theta(\tilde{x}_t, t, s) \|^2\right],
  \label{eq:loss}
\end{align}
where condition $s$ is one style feature in $\mathcal{S}$ derived from the pre-trained conditioner, and it is randomly dropped during the training.

During the sampling phase, a sequence of samples $\tilde{x}_T, \dots, \tilde{x}_0$ is generated starting from $\tilde{x}_T \sim \mathcal{N}(0, \textbf{I})$. 
For each timestep $t$, the model refines the process of denoising $\tilde{x}_{t-1}$ based on $\tilde{x}_t$ through the following operation:
\begin{equation}
  \label{free}
\hat{\epsilon}_{\theta} = \epsilon_\theta(\tilde{x}_t, t, \emptyset) + \omega\big(\epsilon_{\theta}(\tilde{x}_t,t, s) - \epsilon_\theta(\tilde{x}_t, t, \emptyset)\big),
\end{equation}
where $\omega$ is a scalar hyperparameter that controls alignment between the guidance signal and the sample \cite{ho2022classifier}.
Through the iterative application of Eq. (\ref{free}), the diffusion model is capable of sampling new time-series samples that conform to specific styles $s\in \mathcal{S}$. 
It is worth noting that the styles $S_1,\dots,S_{n^s}$ are robust to the class labels, the generation process guarantees the first criterion of domain padding: each generated sample belongs to a known class under the guidance of a single condition.

\subsection{Beyond One Activity Style} \label{multi_sec}
So far our approach has not yet achieved the second criterion of domain padding, as samples conditioned on a singular style $s\in \mathcal{S}$ could demonstrate a limited range of variation within the intra-domain space. Therefore, we then propose a style-fused sampling strategy to further enhance the diversity. This strategy guides the diffusion to generate new data that satisfy any number and combination of styles (conditions), rather than just one. By doing so, the generated data fuse diverse inter and intra-domain styles, effectively meeting the second criterion of domain padding. 

\textbf{Random style combination.}~~
Random style combination entails the ensemble of \emph{one or multiple} style features under a unified class to establish a new diffusion sampling condition. Importantly, the ensemble styles must belong to the same class to preserve class consistency. For each class label $c$, we randomly select any number of style features from the class-specific set $\mathcal{S}^c$ and combine them in all possible ways. This will end up with $2^k-1$ different style combinations (excluding the empty set), where $k=|\mathcal{S}^c|$ is the number of styles in $\mathcal{S}^c$. Mathematically, the collection of all possible style combinations for class $c$ can be expressed by the power set $\mathcal{P}(\mathcal{S}^c)$ of $\mathcal{S}^c$:
\begin{equation}
  \label{eq:power_set_style_combination}
  \mathcal{P}(\mathcal{S}^c) = \{\mathcal{D}_j|\mathcal{D}_j\subseteq \mathcal{S}^c, \mathcal{D}_j\neq \emptyset\}.
\end{equation}
For instance in Fig.~\ref{all} (c), three styles in $\mathcal{S}^c=\{S_1,S_2,S_3\}$ results in 7 different combinations\footnote{$\mathcal{P}(\mathcal{S}^c)=\big\{\{S_1\},\{S_2\},\{S_3\},\{S_1,S_2\},\{S_1,S_3\},\{S_2,S_3\},\{S_1,S_2,S_3\}\big\}$}. This operation is replicated across all classes $1, \ldots, C$, integrating them into a comprehensive style combination set $\mathcal{D}=\{\mathcal{P}(\mathcal{S}^1)\cup\dots\cup \mathcal{P}(\mathcal{S}^C)\}$. The randomness in selecting style combinations can significantly foster diversity within and between domains, and maximize the exploitation of existing styles to generate highly diverse condition space $\mathcal{X}^{\text{cond}}$.

\textbf{Style-fused sampling.}~~
Subsequently, our efforts are directed towards empowering the diffusion model to fuse multiple styles during the data generation conditioned on a specific style combination $\mathcal{D}_j \in \mathcal{D}$. Assuming the diffusion has learned the data distributions $\{\epsilon_\theta(\tilde{x}_t, t, s)\}_{i=1}^{n_s}$ through Eq.~(\ref{eq:loss}), sampling from the composed data distribution $q(\tilde{x}_0|\mathcal{D}_j)$ for any given style combination $\mathcal{D}_j \in \mathcal{D}$ is achieved using the below perturbed noise:
\begin{equation}
  \label{multi}
  \hat{\epsilon}_{\theta}= \epsilon_{\theta}(\tilde{x}_t,  t, \emptyset) + \omega \sum_{s\in \mathcal{D}_j} \Big(\epsilon_{\theta}(\tilde{x}_t,  t,s) - \epsilon_{\theta}(\tilde{x}_t,  t, \emptyset)\Big).
\end{equation}
The derivation of Eq. (\ref{multi}) is provided in Appendix \ref{append:theo}. 
This indicates that while the diffusion training process primarily focuses on an individual style, we can flexibly combine these styles during sampling. 
For instance, consider the combination of $\mathcal{D}_j = \{S_1, S_3\}$ in Fig.~\ref{all} (c) and (d). Each element represents a style associated with the [\textit{walking}] activity. Eq.~(\ref{multi}) can generate new samples with the [\textit{walking}] label possesses unique characteristics that fuse these two styles. This is critical for inter and intra-domain diversity in domain padding: diffusion can flexibly incorporate class-specific instance-level styles from different or the same domains to generate new samples with novel domain distribution. Moreover, given the existence of sub-domains within each domain, our diffusion model is capable of synthesizing novel domains, even from sampling instances within the same domain (we verify this later in the experiments).

\subsection{Workflow of DI2SDiff} \label{detail_sec}
Finally, we elaborate on the comprehensive workflow of our approach, which we refer to as \textbf{D}iversified \textbf{I}ntra- and \textbf{I}nter-domain distributions via activity \textbf{S}tyle-fused \textbf{Diff}usion modeling (DI2SDiff).

\textbf{Architectural design.}~~
The diffsuion model $\epsilon_{\theta}: \tilde{\mathcal{X}^s} \times \mathbb{N} \times \mathcal{X}^{\text{cond}} \rightarrow \tilde{\mathcal{X}^s}$ is built upon a UNet architecture~\cite{ho2022classifier} with repeated convolutional residual blocks. To accommodate the characteristics of time series input, we adapt 2D convolution to 1D temporal convolution. The model incorporates a timestep embedding module and a condition embedding module, each of which is a multi-layer perceptron (MLP). The condition embedding module is used to encode each activity style $s\in \mathcal{S}$, and in the unconditional case $s=\emptyset$, we zero out the entries of $s$. These embeddings are then concatenated and fed into each block of the UNet.

\textbf{Training.}~~
During the training stage, as shown in Fig. \ref{all} (a) and (b), the pre-trained style conditioner extracts style features $\{S_i\}_{i=1}^{n^s}$ for the training instances $\{\mathbf{X_i}\}_{i=1}^{n^s}$. Each data instance $\mathbf{X_i}$, paired with its style $S_i$ and a randomly sampled timestep $t\sim \mathcal{U}$, forms a tripartite input $(\mathbf{X_i}, t, S_i)$. This setup facilitates the optimization of the model against a loss function defined by Eq. (\ref{eq:loss}).

\textbf{Sampling.}~~
During the sampling stage, we construct the style combination set $\mathcal{D}$ by Eq.~(\ref{eq:power_set_style_combination}). As shown in Fig.~\ref{all} (c), a specific style combination $\mathcal{D}_j \in \mathcal{D}$ is then selected to guide the diffusion process, generating the new sample that fuses the styles in $\mathcal{D}_j$. The sampling operates under single-condition guidance when the style combination comprises a single style, i.e., $|\mathcal{D}_j|=1$. Conversely, when the style combination comprises multiple styles, i.e., $|\mathcal{D}_j|>1$, the sampling proceeds under multiple-condition guidance.

\textbf{Domain space diversity.}~~
Through the iterative execution of the sampling procedure, we can generate a diverse range of new, unseen samples that meet domain padding criteria. These synthetic samples collectively form a synthetic dataset $\tilde{D}^s$ for TSC models' training. This process involves two hyperparameters $\kappa$ and $o$. $\kappa$ denotes the proportion of synthetic to original training samples, effectively managing the volume of synthetic samples. $o$ denotes the maximum number of style features that can be combined in each style set. 

\textbf{Training TSC model.}~~
Utilizing the synthetic dataset $\tilde{D}^s$, we are able to augment the training dataset to $\{\tilde{D}^s \cup D^s\} = \{(\mathbf{X}_i, y_i)\}_{i=1}^{n^s + \tilde{n}^s}$. This augmented dataset can be straightforwardly utilized for standard TSC task training. To extract more value from the dataset, we consider the diversity learning strategy from \cite{qin2023generalizable} for the TSC model's training. The main thought goes beyond just minimizing not only the standard cross-entropy loss for correct classification. It also involves minimizing the additional cross-entropy loss to effectively differentiate between synthetic and original samples.
% classifying labels accurately by minimizing standard cross-entropy loss $\mathcal{L}_\text{cls-spe}$, but also to effectively distinguish between synthetic and original samples by minimizing the origin-specific cross-entropy  loss $\mathcal{L}_\text{ori-spe}$ and class-origin joint cross-entropy  loss $\mathcal{L}_\text{cls-ori}$.
For more details, please refer to Appendix~\ref{append:diff} and Appendix~\ref{append:stra}.

% \vspace{-0.5em}

% Our strategy ensures that the classifier retains the essential features learned from the original data while also acquiring novel information from synthetic samples.
\section{EXPERIMENTS}
In this section, we conduct a comprehensive evaluation of DI2SDiff across various cross-person activity recognition tasks to demonstrate \textbf{(1)} its ability to achieve domain padding and significantly diversify the domain space; \textbf{(2)} its outstanding performance in domain generalization; \textbf{(3)} a detailed ablation and sensitivity analysis; and \textbf{(4)} its versatility in boosting existing DG baselines.

\begin{figure}[tbp]    
  \begin{minipage}[t]{0.3\linewidth}
  \centering
  \includegraphics[width=\linewidth]{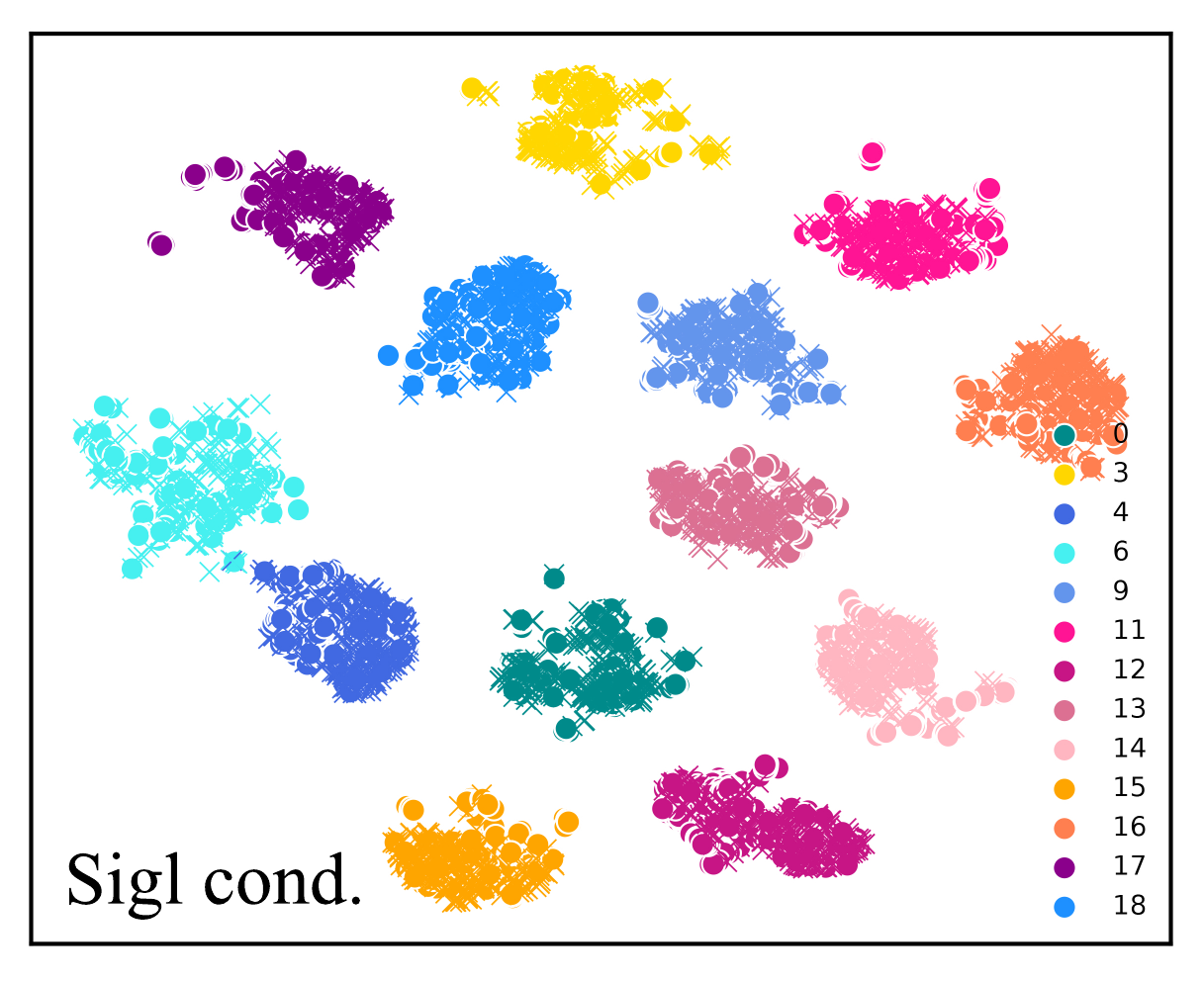}
  \label{fig:subfig4}
  \end{minipage}
  \hfill
  \begin{minipage}[t]{0.3\linewidth}
  \centering
  \includegraphics[width=\linewidth]{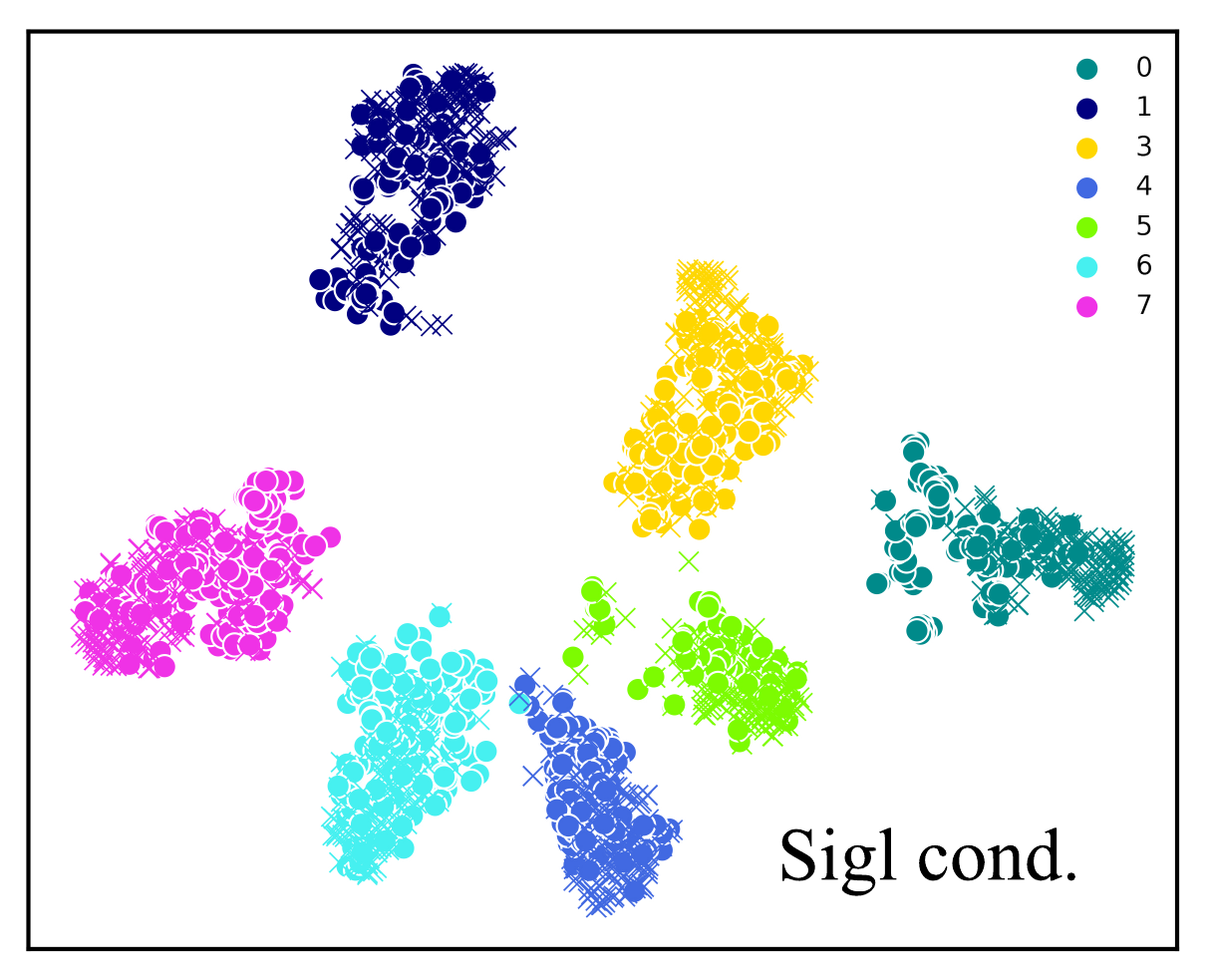}
  \label{fig:subfig5}
  \end{minipage}
  \hfill
  \begin{minipage}[t]{0.3\linewidth}
  \centering
  \includegraphics[width=\linewidth]{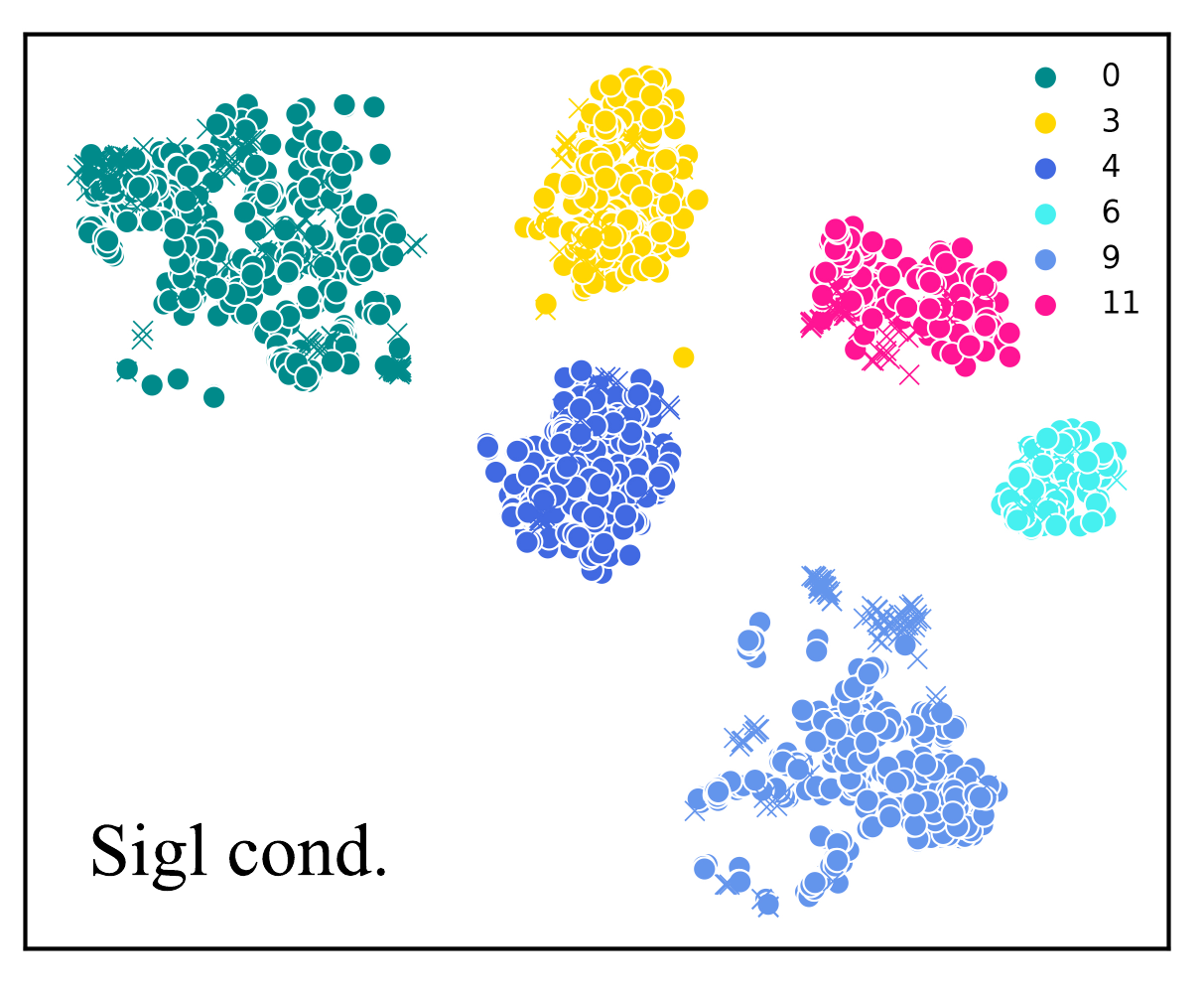}
  \label{fig:subfig6}
  \end{minipage}

   \vspace{-1em}
  
  \begin{minipage}[t]{0.3\linewidth}
  \centering
  \includegraphics[width=\linewidth]{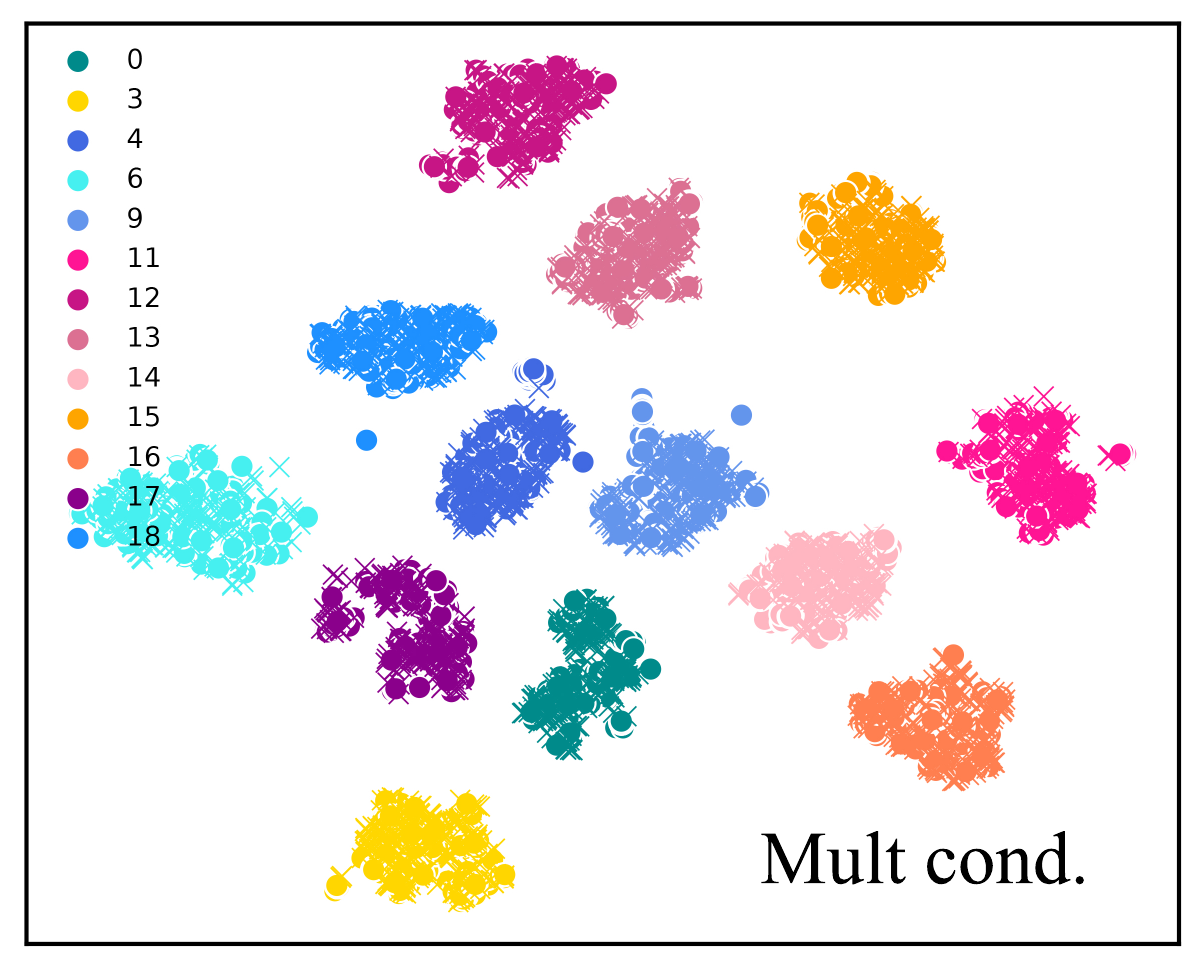}
  \vspace{-2em}
  \caption*{(a) DSADS}
  \label{fig:subfig1}
  \end{minipage}
  \hfill
  \begin{minipage}[t]{0.3\linewidth}
  \centering
  \includegraphics[width=\linewidth]{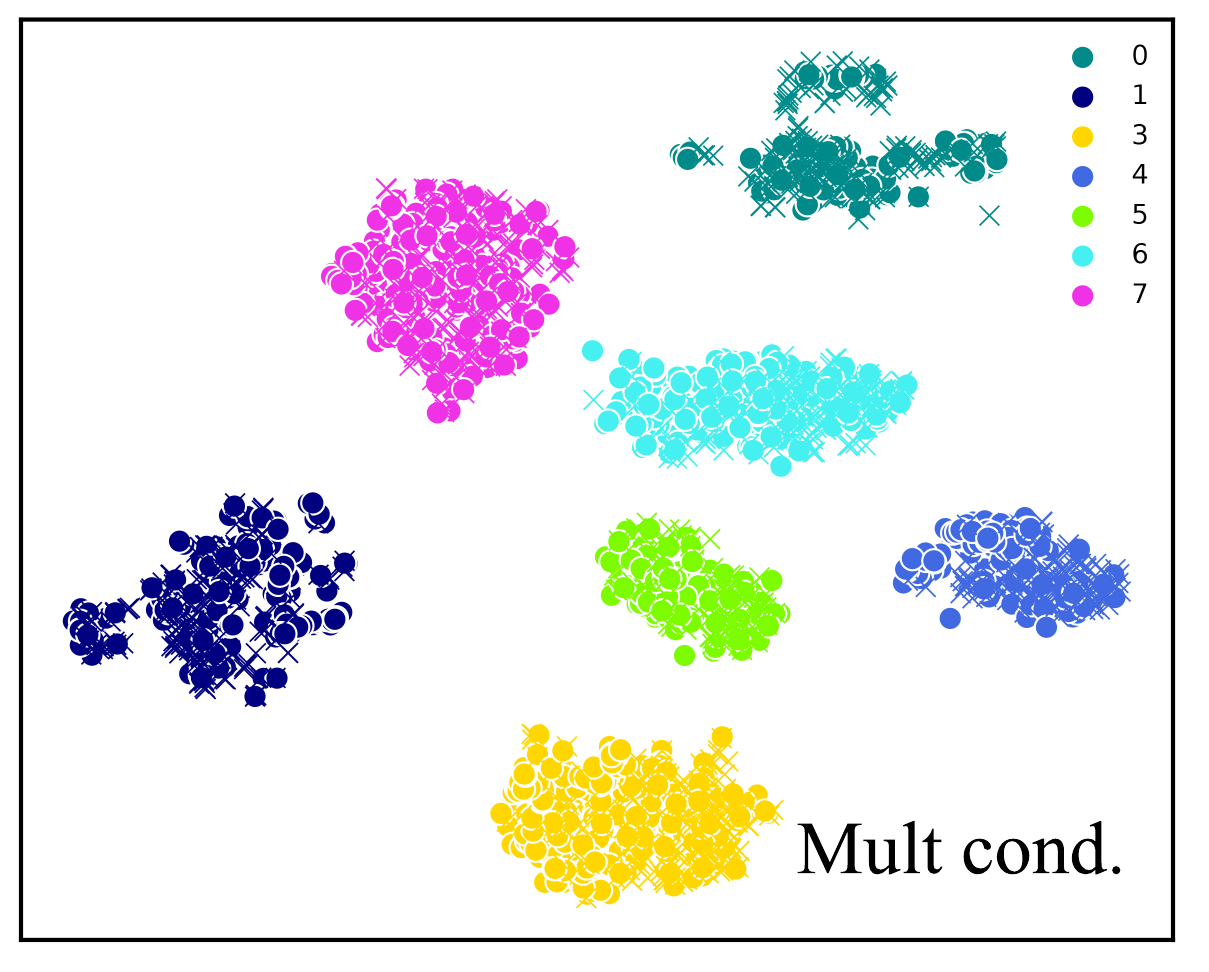}
  \vspace{-2em}
  \caption*{(b) PAMAP2}
  \label{fig:subfig2}
  \end{minipage}
  \hfill
  \begin{minipage}[t]{0.3\linewidth}
  \centering
  \includegraphics[width=\linewidth]{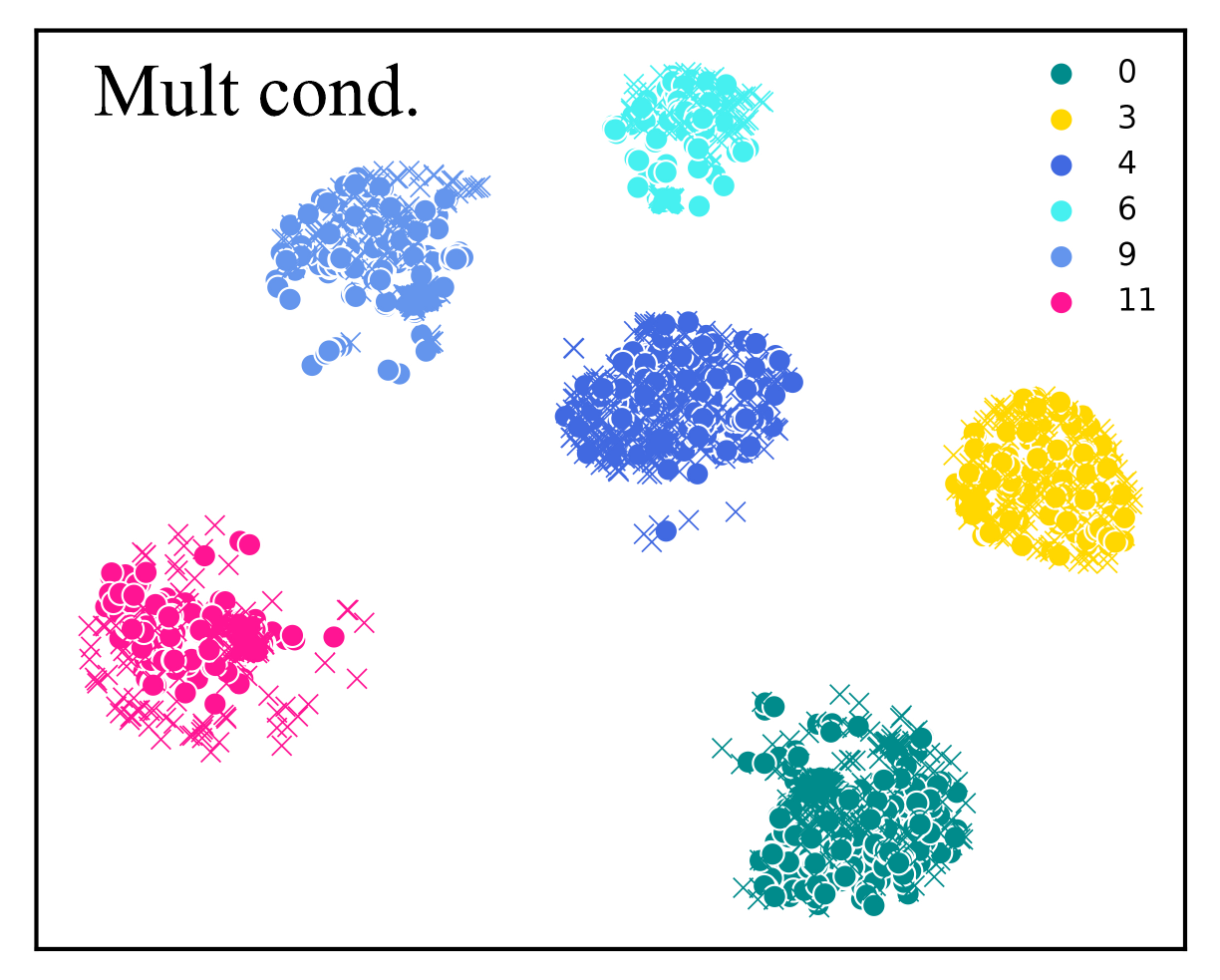}
  \vspace{-2em}
  \caption*{(c) USC-HAD}
  \label{fig:subfig3}
  \end{minipage}
  \vspace{-1em}
  \caption{T-SNE visualization of DSADS, PAMAP2, USC-HAD datasets. Each method generates the same amount of synthetic data. The original and synthetic data are represented by shapes dots and crosses, and each class is denoted by a color. Best viewed in color and zoom in.}
  \label{fig:class-tsne}
  \vspace{-.2em}
\end{figure}

\begin{figure}[t]
  \centering
  \includegraphics[width=1\columnwidth]{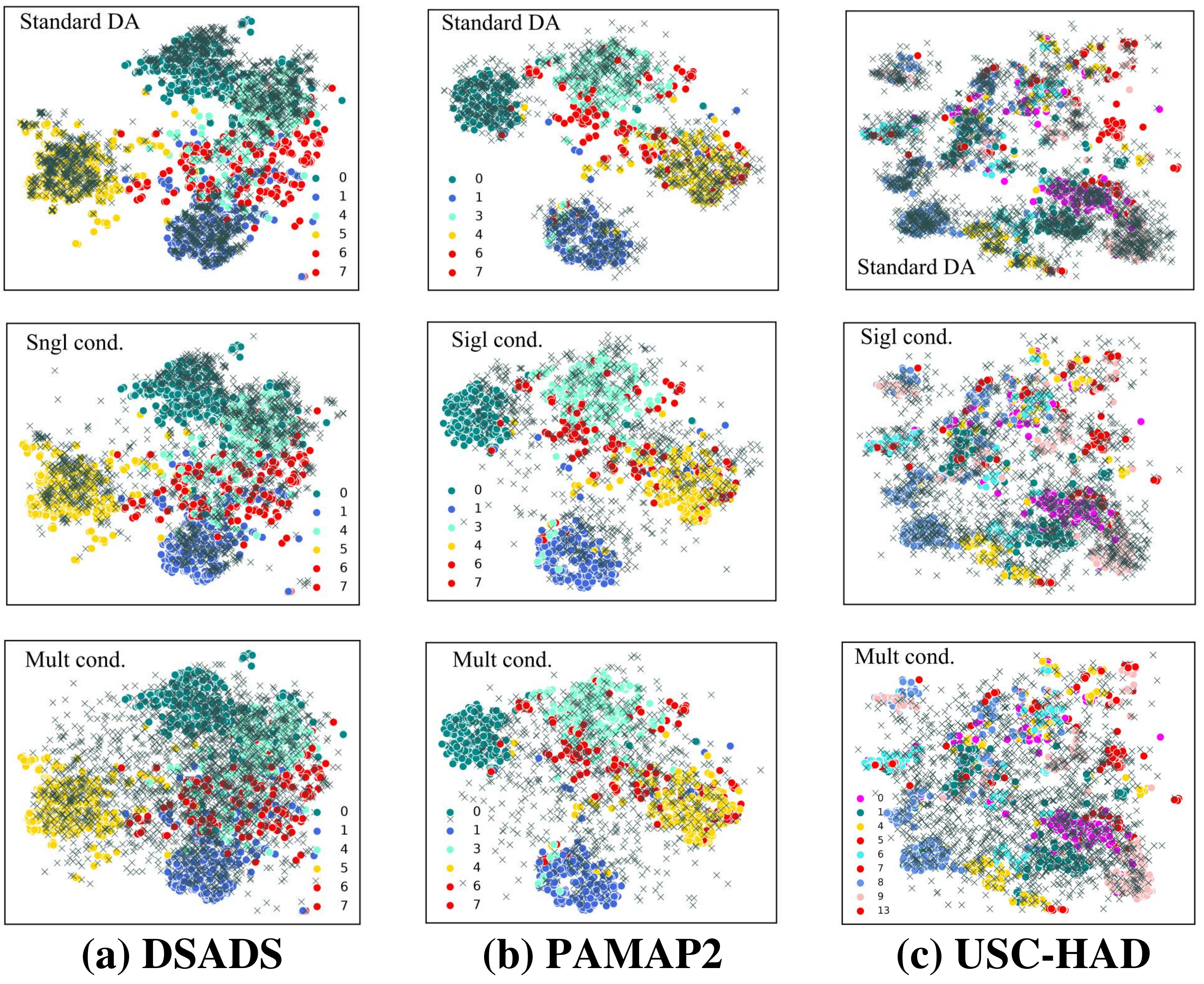}
 
  \caption{T-SNE visualization of DSADS, PAMAP2, and USC-HAR datasets. Each method generates the same amount of synthetic data. Each domain category is represented by a color, and the target domain is represented by a red dot. The original and synthetic data are represented by shapes dots and crosses, respectively. Best viewed in color and zoom in.}
  \label{fig:domain-tsne}
  \vspace{-1.8em}
\end{figure}

% \vspace{-0.5em}
\subsection{Experimental Setup}
\textbf{Datasets.}~~
We assess our method on three widely used HAR datasets: UCI Daily and Sports Dataset (DSADS) \cite{barshan2014recognizing}, PAMAP2 dataset \cite{reiss2012introducing} and USC-HAD dataset \cite{zhang2012usc}. 
We follow the same experimental settings in \cite{qin2023generalizable} that provided a generalizable cross-person scenario. Specifically, the subjects are organized into separate groups for leave-one-out validation.
We assign the data of one group as the target domain and utilize the remaining subjects' data as the source domain. Each subject is treated as an independent task.

\textbf{Baselines.}~~
We compare our approach with a wide range of closely related, strong baselines adapted to TSC tasks. We first select \texttt{Mixup} \cite{xu2020adversarial}, \texttt{RSC} \cite{huang2020self}, \texttt{SimCLR} \cite{chen2020simple}, \texttt{Fish} \cite{shi2021gradient}, and \texttt{DDLearn} \cite{qin2023generalizable}, given their outstanding performance in most recent study~\cite{qin2023generalizable}. Notably, \texttt{DDLearn} \cite{qin2023generalizable} is ranked as the top-performing method. We also include \texttt{TS-TCC} \cite{eldele2021time} for its remarkable generalization performance in self-supervised learning. Additionally, we incorporate \texttt{DANN} \cite{ganin2016domain} and \texttt{mDSDI} \cite{bui2021exploiting}, which are designed to address domain-invariant and domain-specific feature learning, respectively.
In our analysis, the standard data augmentation (DA) techniques \cite{um2017data} are identical to those employed in \cite{qin2023generalizable}, such as scaling and jittering.

\begin{table*}[t]{
  \tiny
  \centering
  \caption{Classification accuracy (\%) ($\pm$ standard deviation) on three public datasets, where each task only comprises 20\% of 
  training data. The best results are marked in bold. ``T0-T'' represent different cross-person activity recognition tasks.}
  \label{tab-acc-all}
  \vspace{-1em}
  \resizebox{.9\textwidth}{!}{%
  \begin{tabular}{lc|ccccccccc}
    \toprule
    & {Tar} & {Mixup~\cite{xu2020adversarial}} & {RSC~\cite{huang2020self}} & {SimCLR~\cite{chen2020simple}} & {Fish~\cite{shi2021gradient}} & {DANN ~\cite{ganin2016domain}} & {mDSDI ~\cite{bui2021exploiting}} & {TS-TCC~\cite{eldele2021time}} & {DDLearn~\cite{qin2023generalizable}} & {Ours} \\\midrule

    \multirow{5}{*}{\rotatebox{90}{DSADS}} & T0 & 74.77 ($\pm1.76$) & 54.32 ($\pm2.19$) & 72.48 ($\pm3.18$) & 55.06 ($\pm1.60$) & 72.49 ($\pm3.21$) & 76.91 ($\pm2.34$) & 80.47 ($\pm 0.53$) &  {87.88 ($\pm1.92$)} & \textbf{{89.93}  ($\pm2.57$)} \\
    & T1 & 75.78 ($\pm3.95$) & 63.62 ($\pm10.56$) & 76.61 ($\pm2.56$) & 62.28 ($\pm3.13$) & 69.61 ($\pm1.96$) & 76.02 ($\pm1.56$) & 79.68 ($\pm 0.42$) &  {88.80 ($\pm1.11$)}  &\textbf{{90.17} ($\pm0.84$)} \\
    & T2 & 74.18 ($\pm4.36$) & 66.48 ($\pm1.80$) & 78.25 ($\pm0.92$) & 68.15 ($\pm1.60$) & 78.97 ($\pm4.06$) & 72.71 ($\pm0.98$) & 84.37 ($\pm 1.87$) &  {89.21 ($\pm1.23$)}  &\textbf{{91.39} ($\pm1.31$)} \\
    & T3 & 75.85 ($\pm3.45$) & 64.29 ($\pm3.37$) & 76.49 ($\pm0.91$) & 68.83 ($\pm3.83$) & 78.54 ($\pm2.14$) & 79.58 ($\pm1.29$) & 82.09 ($\pm 2.51$) &  {85.63 ($\pm1.13$)}  &\textbf{{88.95} ($\pm1.79$)} \\
    & \textbf{Avg} & {75.15} ($\pm2.36$) & {62.18} ($\pm4.32$) & {75.96} ($\pm1.25$) & {63.58} ($\pm0.37$) & {74.90} ($\pm2.63$) & {76.31} ($\pm1.56$) &{81.65} ($\pm 1.33$) &  {{87.88} ($\pm0.82$)}  &\textbf{{90.11} ($\pm1.63$)} \\\midrule
   \multirow{5}{*}{\rotatebox{90}{PAMAP2}} & T0 & 57.81 ($\pm0.55$) & 55.99 ($\pm1.29$) & 63.28 ($\pm3.33$) & 54.04 ($\pm4.31$) & 54.02 ($\pm3.52$) & 58.70 ($\pm3.14$) & 64.08 ($\pm 1.98$) &  {75.55 ($\pm0.79$)}  &\textbf{{79.58} ($\pm2.46$)} \\
    & T1 & 81.51 ($\pm3.94$) & 83.08 ($\pm2.42$) & 81.25 ($\pm1.59$) & 85.16 ($\pm1.39$) & 77.21 ($\pm3.79$) & 83.82 ($\pm1.62$) & 86.55 ($\pm2.28$) &  {90.07 ($\pm2.40$)}  &\textbf{{94.12} ($\pm1.20$)} \\
    & T2 & 77.34 ($\pm3.33$) & 78.65 ($\pm3.99$) & 78.65 ($\pm1.87$) & 79.69 ($\pm4.00$) & 78.80 ($\pm1.87$) & 79.15 ($\pm2.72$) & 80.21 ($\pm 0.52$) &  {85.51 ($\pm0.76$)}  &\textbf{{89.57} ($\pm2.48$)} \\
    & T3 & 70.31 ($\pm5.64$) & 68.10 ($\pm6.27$) & 71.09 ($\pm1.99$) & 72.53 ($\pm0.49$) & 61.96 ($\pm2.11$) & 78.61 ($\pm0.49$) & 77.32 ($\pm 0.47$) &  {80.67 ($\pm1.78$)}  &\textbf{{84.75} ($\pm3.72$)} \\
    & \textbf{Avg} & {71.74} ($\pm1.37$) & {71.45} ($\pm2.55$) & {73.57} ($\pm1.21$) & {72.85} ($\pm0.37$) &{68.00} ($\pm2.66$) & {75.07} ($\pm1.99$) & {77.04} ($\pm1.29$) &  {{82.95} ($\pm0.60$)}  &\textbf{{87.01} ($\pm1.94$)} \\\midrule
   \multirow{6}{*}{\rotatebox{90}{USC-HAD}} & T0 & 68.66 ($\pm4.67$) & 75.69 ($\pm4.28$) & 69.36 ($\pm2.34$) & 73.70 ($\pm3.97$) & 57.79 ($\pm4.73$) & 59.71 ($\pm 1.23$) & 78.96 ($\pm0.79$) &  {79.06 ($\pm2.11$)}  &\textbf{{88.33} ($\pm1.70$)} \\
    & T1 & 68.75 ($\pm1.29$) & 72.40 ($\pm2.88$) & 66.62 ($\pm1.44$) & 72.05 ($\pm2.93$) & 64.95 ($\pm2.68$) & 67.35 ($\pm 2.46$) & 79.55 ($\pm1.23$) &  {80.15 ($\pm1.11$)}  &\textbf{{81.64} ($\pm0.28$)} \\
    & T2 & 71.79 ($\pm0.65$) & 72.83 ($\pm3.62$) & 76.04 ($\pm1.61$) & 69.10 ($\pm2.93$) & 71.97 ($\pm3.23$) & 63.89 ($\pm 3.69$) & 78.15 ($\pm2.15$) &  {80.81 ($\pm0.74$)} & \textbf{{88.37} ($\pm1.46$)} \\
    & T3 & 61.29 ($\pm3.90$) & 63.19 ($\pm5.30$) & 61.24 ($\pm1.06$) & 58.51 ($\pm3.66$) & 45.65 ($\pm2.18$) & 63.87 ($\pm 4.92$) & 64.35 ($\pm1.58$) &  {70.93 ($\pm1.87$)} & \textbf{{77.84} ($\pm1.10$)} \\
    & T4 & 65.63 ($\pm4.55$) & 66.75 ($\pm3.25$) & 62.85 ($\pm2.17$) & 63.72 ($\pm8.31$) & 54.94 ($\pm3.56$) & 55.95 ($\pm 6.15$) & 70.25 ($\pm0.88$) &  {75.87 ($\pm2.99$)}  &\textbf{{83.84} ($\pm0.88$)} \\
    & \textbf{Avg} & {67.22} ($\pm2.41$) & {70.17} ($\pm3.51$) & {67.22} ($\pm0.39$) & {67.42} ($\pm3.91$) & {59.06} ($\pm2.65$) & {62.15} ($\pm 3.08$) & {74.25} ($\pm1.16$) &  {{77.36} ($\pm0.99$)}  &\textbf{{84.00} ($\pm1.09$)} \\\hline
   \multicolumn{2}{c|}{\textbf{Avg All}} & {71.37} & {67.93} & {72.25} & {67.95} & {67.32} & {71.18} & {77.65} &  {82.73}  &\textbf{{87.04}} \\\hline
  \end{tabular}}
  }
  % \vspace{-0.7em}
\end{table*}

\textbf{Architecture and implementation.}
For fairness, we adopt the same feature extractor as described in~\cite{qin2023generalizable}, which consists of two blocks for DSADS and PAMAP2, and three blocks for USC-HAD. Each block includes a convolution layer, a pooling layer, and a batch normalization layer. All baselines, except TS-TCC \cite{eldele2021time}, employ this feature extractor. In each experiment, we report the average performance and standard deviation over three random seeds.
For detailed experimental setups, including dataset details and training procedures, please refer to Appendix \ref{append:exp}.

\vspace{-1.em}
\subsection{Domain Padding and Diversity Evaluation}
In this part, we demonstrate whether our approach can effectively diversify the domain space and generate diverse samples that meet domain padding criteria. To this end, we adopt T-SNE \cite{van2008visualizing} to visualize the latent feature space in terms of class and domain dimensions.

% \noindent
\textbf{(1) Class-Preserved Generation.} 
Firstly, we evaluate the class consistency of synthetic data, i.e., the first criterion of domain padding. We employ a class feature extractor, trained with class labels, to map both original and synthetic data into a class-specific latent space. The results of single-condition guidance ($|\mathcal{D}_j|=1$) and multiple-condition guidance ($|\mathcal{D}_j|>1$) are shown in Fig. \ref{fig:class-tsne}.

It can be observed that all synthetic samples (crosses) are closely clustered around their corresponding original instances and classes (dots). This clustering indicates that our method effectively maintains class information, avoiding the introduction of class noise; importantly, this holds true under both single and multiple-condition guidance. Moreover, the use of multiple-condition guidances appears to enhance class discriminability more than single-condition guidance in Fig. \ref{fig:class-tsne}. This enhancement is likely because more guidance signals provide more robust class semantics (akin to ensemble learning), therefore resulting in a better class alignment.

% \noindent
\textbf{(2) Intra- and Inter-Domain Diversity. }
Next, we evaluate the intra- and inter-domain diversity of the synthetic data, i.e., the second criterion of domain padding. We train the domain feature extractor on source domains with domain labels. We then map source and target data into a domain-specific latent space, and compare the synthetic data from the standard DA method, single-condition guidance ($|\mathcal{D}_j|=1$), and multiple-condition guidance ($|\mathcal{D}_j|>1$). The results are shown in Fig. \ref{fig:domain-tsne}.

The findings reveal that the standard DA method generates tightly clustered samples (crosses) around the original data (dots), falling short of diversifying the domain space, particularly the inter-domain space. Our single-condition guidance method offers a partial solution and generates sparse data between different domains thanks to diffusion's probabilistic nature. However, relying solely on a single-style guidance approach has limitations for domain padding.
The introduction of our style combinations in Eq. (\ref{eq:power_set_style_combination}) makes a substantial improvement: the multiple-condition guidance excels in ``padding'' the distributional gaps both within and between source domains, as demonstrated in Fig. ~\ref{fig:domain-tsne}.

Moreover, as indicated in Fig. \ref{fig:domain-tsne}, through multiple-condition guidance, the synthetic samples (crosses) closely resemble the target domain data (red dots) and demonstrate less dependence on the specific characteristics of source domains. This demonstrates that the fusion of multiple style features creates a new style. This is of importance for the domain generalization in TSC. Given the significant differences in individual styles and the small-scale nature of source domain data, our style-fused sampling demonstrates great potential to simulate various new and unseen distributions, from which the TSC model can better adapt to the target domains.

In addition, we can observe in Fig. \ref{fig:domain-tsne} (c) that the USC-HAD dataset presents an additional challenge of intra-domain gaps due to its fragmented and sparsely distributed source domains with distinct sub-domains. These gaps contribute to an increased distribution shift, posing difficulties for existing DG methods to perform effectively (We show their results in Tab. \ref{tab-acc-all} in later). Through random instance-level style fusion, our approach effectively addresses this sub-domain challenge, enabling the synthesis of new data distribution within sub-domains. As a result, our method can yield exceptional performance on the complex tasks like USC-HAD.

\vspace{-1em}
\subsection{Generalization Performance}
Now we conduct a series of experiments to evaluate the generalization performance of DI2SDiff against other strong DG baselines.

\textbf{Overall performance.} Tab. \ref{tab-acc-all} presents a comparative analysis of the classification accuracies achieved by all DG methods across three datasets, each task of which comprises 20\% of the training data. As we can see, representation learning baselines that focus solely on learning domain-invariant features, such as DANN~\cite{ganin2016domain}, exhibit suboptimal performance due to the limited diversity of the training data in HAR. The method mDSDI~\cite{bui2021exploiting}, on the other hand, achieves improved performance by additionally learning domain-specific features. However, it does not match the performance of DDLearn~\cite{qin2023generalizable}, which utilizes data augmentation, underscoring the importance of training data diversity in enhancing generalization in HAR. In contrast, our DI2SDiff, leveraging advanced synthesis of highly diverse data across both intra- and inter-domain space, markedly surpasses all baseline methods in every task. 
In addition, we observe that all baselines, including DDLearn, demonstrate poor performance on the USC-HAD dataset.
As we discussed in Fig. \ref{fig:domain-tsne} (c), this decline is due to the presence of sub-domains within the source domain, which poses a highly
challenging problem in DG. Nevertheless, DI2SDiff adeptly addresses this issue by integrating instance-level style fusion, thereby synthesizing new data distributions between the sub-domains. As a result, our approach achieves outstanding performance, outperforming the second-best method by a clear margin (6.64\%) in USC-HAD.

\begin{table}[t]
  \centering
  \small
  \caption{Classification accuracy (\%) on three public datasets with varying percentages (\%) of used training data.}
  \label{tab-rate}
  \vspace{-1em}
  \resizebox{.48\textwidth}{!}{%
  \begin{tabular}{@{\,\,\,\,}l@{\,\,\,\,}c@{\,\,\,\,}|@{\,\,\,\,}c@{\,\,\,\,}c@{\,\,\,\,}c@{\,\,\,\,}c@{\,\,\,\,}c@{\,\,\,\,}c@{\,\,\,\,}c@{\,\,\,\,}c@{\,\,\,\,}c@{\,\,\,\,}}
  \toprule
  & {Perct.}& {Mixup} & {RSC} & {SimCLR} & {Fish} & {DANN} & {mDSDI} & {TS-TCC} & {DDLearn} & {Ours} \\\midrule
  \multirow{6}{*}{\rotatebox{90}{DSADS}}
  & 20\% & 75.15 & 62.18 & 75.96 & 63.58 & 74.90 & 76.31 & 81.65 & 87.88 & \textbf{{90.11}} \\
  & 40\% & 82.48 & 67.70 & 75.76 & 65.82 & 75.45 & 76.55 & 82.54 & 89.71 & \textbf{{91.25}} \\
  & 60\% & 82.70 & 69.98 & 75.61 & 67.65 & 76.55 & 77.89 & 83.78 & 90.43 & \textbf{{92.56}} \\
  & 80\% & 81.58 & 75.37 & 74.69 & 66.03 & 76.89 & 79.25 & 84.12 & 90.97 & \textbf{{94.58}} \\
  & 100\% & 83.44 & 75.58 & 76.22 & 69.35 & 80.52 & 79.58 & 86.57 & 91.95 & \textbf{{95.23}} \\
  & \textbf{Avg} & {81.07} &{70.16} & {75.65} & {66.49} & {76.86} & {77.92} &{83.73} & {90.19} & \textbf{{92.75}} \\
  \midrule
  \multirow{6}{*}{\rotatebox{90}{PAMAP2}}
  & 20\% & 71.74 & 71.45 & 73.57 & 72.85 & 68.00 & 75.07 & 77.04 & 82.95 & \textbf{{87.01}} \\
  & 40\% & 76.69 & 73.73 & 74.25 & 77.02 & 69.85 & 72.55 & 78.35 & 84.34 & \textbf{{87.66}} \\
  & 60\% & 77.83 & 75.72 & 74.71 & 76.04 & 70.88 & 76.56 & 80.15 & 85.03 & \textbf{{88.75}} \\
  & 80\% & 78.00 & 76.17 & 74.09 & 75.13 & 77.82 & 77.53 & 81.78 & 86.67 & \textbf{{89.92}} \\
  & 100\% & 79.72 & 77.96 & 74.25 & 75.49 & 79.56 & 78.83 & 83.45 & 86.31 & \textbf{{90.96}} \\
  & \textbf{Avg} & {76.80} & {75.01} & {74.17} & {75.31} & {73.22} & {76.11} & {80.15} & {85.06} & \textbf{{88.86}} \\
  \midrule
  \multirow{6}{*}{\rotatebox{90}{USC-HAD}}
  & 20\% & 67.22 & 70.17 & 67.22 & 67.42 & 59.06 & 62.15 & 74.25 & 77.36 & \textbf{{84.00}} \\
  & 40\% & 75.30 & 77.31 & 69.16 & 73.54 & 61.52 & 68.85 & 75.32 & 80.72 & \textbf{{84.97}} \\
  & 60\% & 78.14 & 77.59 & 71.38 & 76.09 & 68.71 & 76.75 & 77.84 & 80.88 & \textbf{{87.53}} \\
  & 80\% & 79.76 & 78.65 & 71.99 & 77.21 & 68.52 & 77.72 & 78.91 & 82.49 & \textbf{{89.25}} \\
  & 100\% & 81.27 & 79.41 & 72.14 & 78.92 & 72.05 & 78.59 & 79.15 & 82.51 & \textbf{{91.13}} \\
  & \textbf{Avg} & {76.34} & {76.62} & {70.38} & {74.64} & {65.97} & {72.81} &{77.09} &{80.80} & \textbf{{87.38}} \\
  \bottomrule
  \end{tabular}
  }
  \vspace{-1.5em}
\end{table}

\textbf{Data proportion analysis.}
In Tab. \ref{tab-rate}, we assess DI2SDiff's performance over a range of data volumes by adjusting the proportion of training data from 20\% to 100\%. The results demonstrate DI2SDiff's consistent superiority over the baseline methods across various proportions of training data. This highlights the ability of our approach to efficiently generate informative samples from varying amounts of available data and effectively learn from them.
As the size of the training sample increases, the advantage of our method becomes more pronounced. For instance, as we increase the size from 20\% to 100\% of USC-HAD, the accuracy improvement grows from 6.64\% to 8.62\% compared to the second-best baseline (DDLearn). This is because the number of style combinations increases exponentially ($2^k-1$) with larger training data volumes, as shown in Eq.~(\ref{eq:power_set_style_combination}). Hence, enlarging the training dataset can provide significant diversity enhancement of data synthesis, leading to more substantial gains in the model's generalization ability.

\begin{table}[tbp]
  \centering
  
  \caption{The results of the ablation study on three datasets and each task is averaged for an overall assessment. }
  \label{tab:ablation} 
  \vspace{-1em}
  \tiny
  \resizebox{.48\textwidth}{!}{%
  \begin{tabular}{@{\,\,}l@{\,\,}|cccccc@{\,\,}}
    \toprule
    \multicolumn{1}{c|}{\multirow{2}[4]{*}{Variants}} & \multicolumn{2}{c}{DSADS} & \multicolumn{2}{c}{PAMAP2} & \multicolumn{2}{c}{USC-HAD} \\
\cmidrule{2-7}      & 20\% & 100\% & 20\%& 100\% & 20\% & 100\% \\
    \midrule
    \makecell[c]{Standard DA}         & 75.58 & 82.57 & 70.31 & 86.41 & 69.14 & 75.45 \\
    \makecell[c]{Class Label Guidance} & 76.25 & 84.67 & 72.78 & 88.52 & 70.85 & 76.26 \\
    \makecell[c]{Single Style Sampling} & 86.98 & 91.12 & 83.07 & 89.64 & 75.27 & 83.57 \\
    \makecell[c]{Style-Fused Sampling} & \textbf{90.11} & \textbf{95.23} & \textbf{87.01} & \textbf{90.96} & \textbf{84.00} & \textbf{91.13} \\
    % \midrule
    % \makecell[c]{$\mathcal{L}$ w/o $\mathcal{L}_{\text{cls-ori}}$ w/o $\mathcal{L}_{\text{ori-spe}}$ } & 87.95 & 91.69 & 83.25 & 88.63 & 81.52 & 88.33 \\
    % \makecell[c]{$\mathcal{L}$ w/o $\mathcal{L}_{\text{cls-ori}}$ } & 88.27 & 92.52 & 84.53 & 89.15 & 81.87 & 88.65 \\
    % \makecell[c]{$\mathcal{L}$ w/o $\mathcal{L}_{\text{ori-spe}}$ } & 89.76 & 93.62 & 86.81 & 90.53 & 83.45 & 90.17 \\
    %  Diversity learning strategy (Ours)& \textbf{90.11} & \textbf{95.23} & \textbf{87.01} & \textbf{90.96} & \textbf{84.00} & \textbf{91.13} \\
    \bottomrule
  \end{tabular} 
  }
  % \vspace{-2em}
\end{table}

\subsection{Ablation and Sensitivity Analysis}
% Table generated by Excel2LaTeX from sheet 'Sheet3'
In this section, we perform an ablation study that focuses on the main step of DI2SDiff, i.e., generating diverse time-series data via diffusion for data augmentation. We keep the number of synthetic samples and the training strategy of TSC models the same for all variants. Additionally, we conduct a sensitivity analysis that focuses on its two hyperparameters: $o$, which controls the maximum number of style features in each style combination, and $\kappa$, which controls the volume of synthetic data.

\begin{figure}[tbp]
  \begin{minipage}[t]{0.45\linewidth}
  \centering
  \includegraphics[width=\linewidth,height=0.60\linewidth]{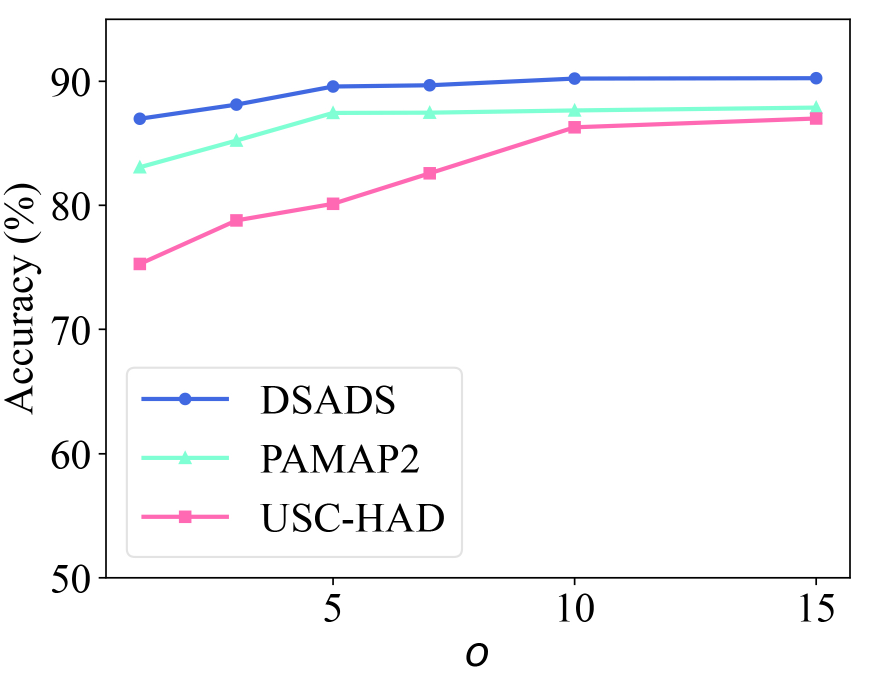}
  \vspace{-2.5em}
  \caption*{\small (a) $o$}   
  \end{minipage}
  \hfill
  \begin{minipage}[t]{0.45\linewidth}
  \centering
  \includegraphics[width=\linewidth,height=0.60\linewidth]{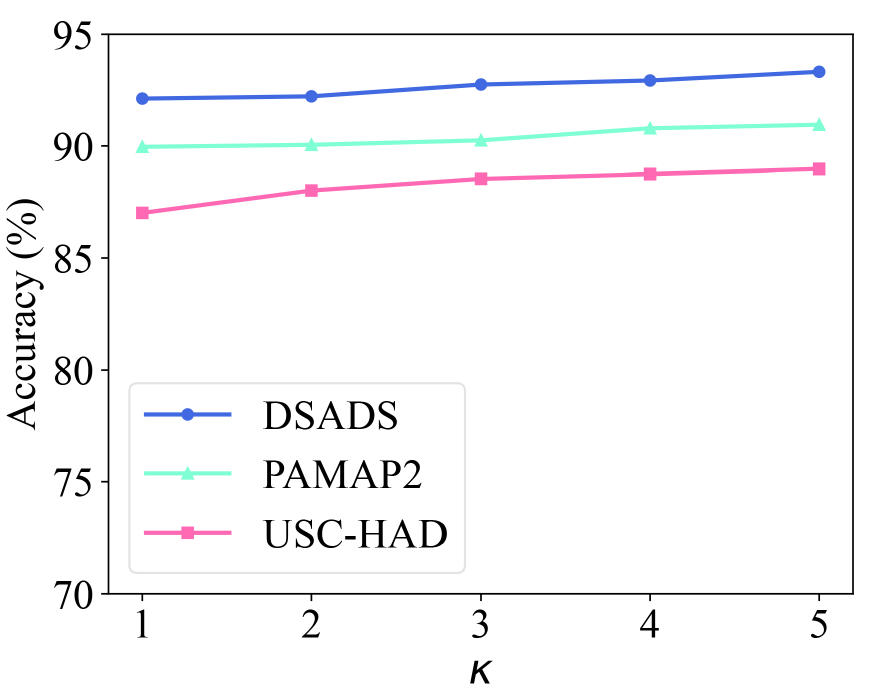}
  \vspace{-2.5em}
  \caption*{\small (b) $\kappa$}
  \end{minipage}
  \hfill

  \vspace{-1em}
  \caption{Hyperparameter sensitivity analysis on $o$ and $\kappa$.}
  % (c) weight of the second loss module $\gamma$, and
  % (d) weight of the third loss module $\eta$.}
  \label{fig:subfig}
  \vspace{-1.1em}
\end{figure}

\textbf{Ablation on diffusion model.}~~
Our findings, as presented in Table \ref{tab:ablation}, indicate that the standard DA method and class label guidance\footnote{This involves directly using the class labels, rather than the style features, as the condition to guide diffusion.} falter in performance. The failure of the class label guidance sampling suggests that using class labels alone, without instance-related information, cannot generate high-quality data. In contrast, the diffusion model that utilizes a single style feature as a condition achieves better performance, suggesting that leveraging representation features for instance-specific sampling can boost the quality of generated data. Moreover, the incorporation of style-fused sampling can further improve generalization by producing samples with distinct features.

\textbf{Hyperparameter sensitive analysis.} We analyze the sensitivity of our hyperparameters by varying one parameter while maintaining the others constant. As shown in Fig. \ref{fig:subfig}, increasing the complexity of style combinations ($o$) and the volume of synthetic data ($\kappa$) generally leads to performance improvement. It becomes non-sensitive when the value is too large. We find that an $o$ value of 5 for the DSADS and PAMAP2 and an $o$ value of 10 for the USC-HAD sufficiently ensure a diverse range of styles. $\kappa$ values of 1 or 2 strike an effective balance between accuracy and training overhead for all three datasets. By flexibly tuning these hyperparameters, we can achieve even greater performance improvements for the TSC model across various tasks while meeting specific needs.

\begin{figure}[tbp]
  \centering
  \includegraphics[width=0.8\linewidth]{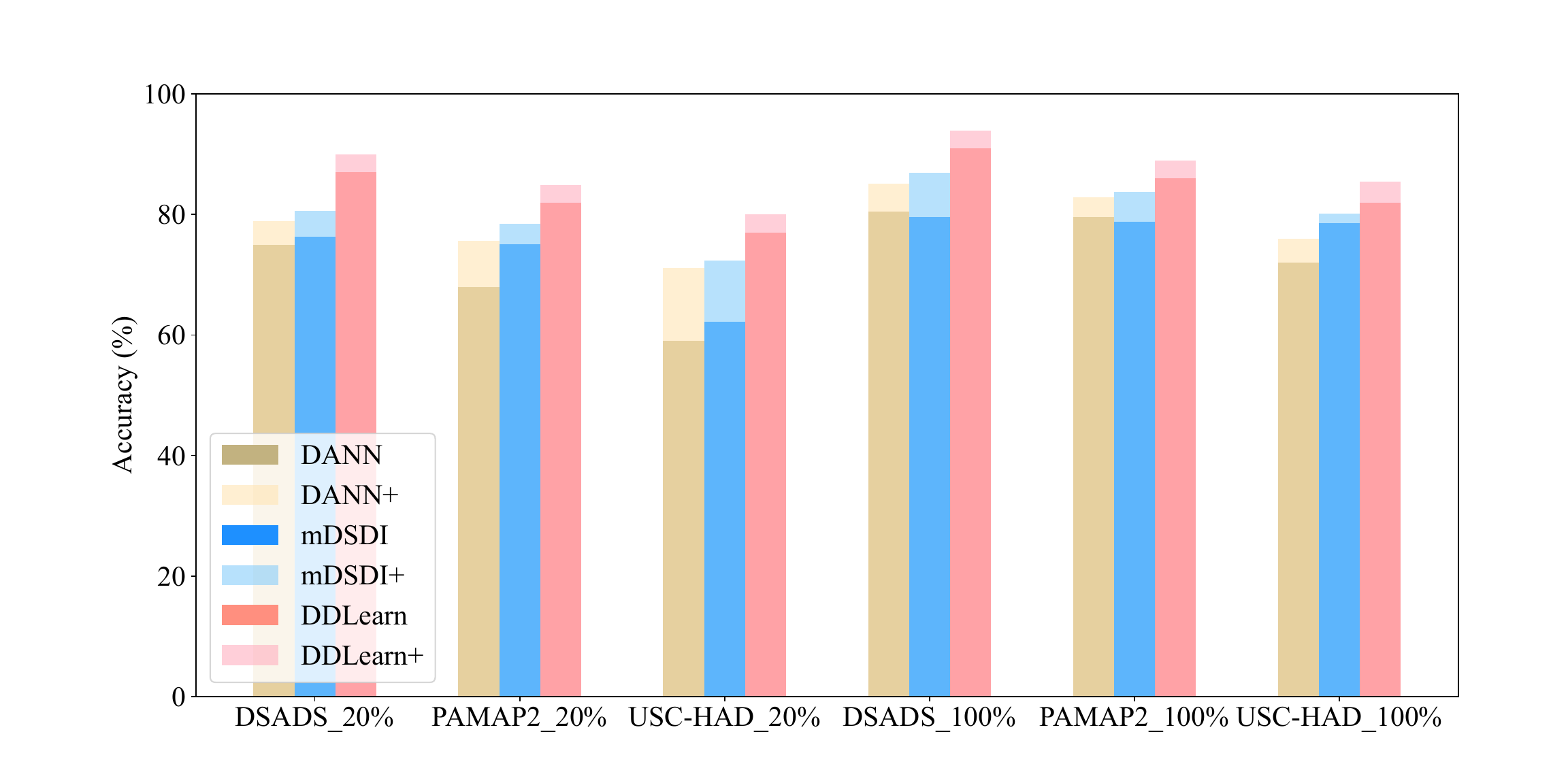}
  \vspace{-1.5em}
  \caption{Enhancing the performance of DANN \cite{ganin2016domain}, mDSDI \cite{bui2021exploiting} and DDLearn \cite{qin2023generalizable} with our data generation (+) on 20\% and 100\% training data in three datasets.
  }
  \label{fig:imp}
  % \vspace{-1em}
\end{figure}

\subsection{Benifits to other DG baselines} 
We demonstrate the versatility of our approach in boosting the performance of existing DG baselines. The results are shown in Fig. \ref{fig:imp}. By incorporating our synthetic data into the training datasets of baselines, we consistently observe performance improvements across the board, including DANN \cite{ganin2016domain}, mDSDI \cite{bui2021exploiting}\footnote{In mDSDI, our synthetic data is treated as a new domain.}, and DDLearn \cite{qin2023generalizable}\footnote{In DDLearn, our synthetic data is treated as a new augmentation method.}.
This demonstrates the versatility of integrating our method to provide additional gains, making it a practical solution for immediate application. The diverse synthetic data of DI2SDiff is thus ready for use, offering a straightforward way to bolster various baselines without necessitating further data generation.

\section{Conclusion}

In this paper, we tackle the key issue of DG in cross-person activity recognition, i.e., the limited diversity in source domain. We introduce a novel concept called ``domain padding'' and propose DI2SDiff to realize this concept. Our approach generates highly diverse inter- and intra-domain data distributions by utilizing random style fusion. Through extensive experimental analyses, we demonstrate that our generated samples effectively pad domain gaps. By leveraging these new samples, our DI2SDiff outperforms advanced DG methods in all HAR tasks. A notable advantage of our work is its efficient generation of diverse data from a limited number of labeled samples. This potential enables DI2SDiff to provide data-driven solutions to various models, thereby reducing the dependence on costly human data collection.

\section*{ACKNOWLEDGEMENTS} 
This work is supported by Zhejiang Science and Technology Plan Project under Grant 2023C03183, Key Scientific Research Base for Digital Conservation of Cave Temples (Zhejiang University), China NSFC under Grant 62172284, and Guangdong Basic and Applied Basic Research Foundation under Grant 2022A1515010155.

% \newpage
\bibliographystyle{ACM-Reference-Format}
\balance
\bibliography{sample-base}

% %%
% %% If your work has an appendix, this is the place to put it.
\newpage
 
\appendix
\section{Details of Style Conditioner}
\label{append:sty}
Our style conditioner is adapted from the contrastive module called Time-Series representation learning framework via Temporal and Contextual Contrasting (TS-TCC), proposed in \cite{eldele2021time}. TS-TCC has demonstrated robust representation learning capability for the HAR task, resulting in each output containing both class-maintained information and unique features of the corresponding time-series instance. This makes it highly suitable for extracting distinctive activity styles from each activity sample. We implemented the model using the official code (\url{https://github.com/emadeldeen24/TS-TCC/}). Here is a brief introduction to the model and how to adjust it for our tasks.

\textbf{Architecture.} TS-TCC \cite{eldele2021time} consists of two components: a feature encoder denoted as $f_{\text{enc}}$ and a Transformer denoted as $f_\text{trans}$. The feature encoder $f_\text{enc}$ is a 3-block convolutional architecture. Each block comprises a convolutional layer, a batch normalization layer, and a ReLU activation function. The Transformer $f_\text{trans}$ primarily consists of successive blocks of multi-headed attention (MHA) followed by an MLP block. The MHA block employs 8 attention heads, and the MLP block is composed of two fully-connected layers with a non-linearity ReLU function and dropout in between. The Transformer stacks  $L$ layers to generate the final features, where $L$ is typically set to 4.

\textbf{Contrastive learning pipeline .} TS-TCC \cite{eldele2021time} uses strong and weak data augmentation techniques to enable contrastive learning of the feature encoder and Transformer. These generate two views for temporal and contextual contrasting, which minimize distance and pull views closer together. Thus, the self-supervised loss combines the temporal and contextual contrastive losses to encourage discriminative representations.

\textbf{Adaptation for our style conditioner.} We adjust the input channel to match the input channel of our training data and set the kernel size to 9. Other components remain consistent with the experimental settings used in the HAR dataset. We maintain the original training settings, such as setting the epoch to 40 and using a Adam optimizer with a learning rate of 3e-4. Once the module is trained, the trained encoder and Transformer are combined to form our style conditioner. When extracting the style from the corresponding original data $\mathbf{X}_i$, the style conditioner produces the representation $Z_i = f_{enc}(\mathbf{X}_i)$ and then outputs the corresponding context vector $S_i = f_{trans}(Z_i) \in \mathbb{R}^H$, where $H$ is its length.

\section{Proof}
\label{append:theo}
We elaborate on how the conditional diffusion model trained with single style condition $\{S_i\}_{i=1}^{n_s}$ fuse multiple style conditions.
From the derivations in~\citep{luo2022understanding, song2020denoising}, we have 
\begin{equation}
  \nabla_{\tilde{x}_t} \log q(\tilde{x}_t|s) \propto -\epsilon_\theta(\tilde{x}_t, t,s).
  \label{eqn:grad}
\end{equation}
Hence, single style conditional data distribution $\{q(\tilde{x}_t|S_i)\}_{i=1}^{n_s}$ can be modeled with a singular denoising model $\{\epsilon_\theta(\tilde{x}_t, t, S_i)\}_{i=1}^{n_s}$ that conditions on the respective style $S_i$.

To integrate multiple styles in a specific sytle combination $\mathcal{D}_j\in \mathcal{D}$, we aim is to model data distribution $q(\tilde{x}_t|\mathcal{D}_j)=q(\tilde{x}_t|\{S_i\}_{i=1}^L)$. Here, for convenience, we use $L$ to denote the number of styles in $\mathcal{D}_j$ and use $\{S_i\}_{i=1}^L$ to denote all styles in $\mathcal{D}_j$. We assume that $\{S_i\}_{i=1}^L$ are conditionally independent given $\tilde{x}_t$. Thus, it can  be factorized as follows:
\begin{align*}
  &q(\tilde{x}_t|\{S_i\}_{i=1}^L) \propto q(\tilde{x}_t) \prod_{i=1}^L \frac{q(\tilde{x}_t|S_i)}{q(\tilde{x}_t)} \quad \text{(Bayes Rule)}\\
  \Rightarrow\; &\log q(\tilde{x}_t|\{S_i\}_{i=1}^L) \propto \log q(\tilde{x}_t) + \sum_{i=1}^L (\log q(\tilde{x}_t|S_i) - \log q(\tilde{x}_t)) \\
  \Rightarrow\; &\nabla_{\tilde{x}_t} \log q(\tilde{x}_t|\{S_i\}_{i=1}^L) = \nabla_{\tilde{x}_t} \log q(\tilde{x}_t) \\
  &\qquad + \sum_{i=1}^L (\nabla_{\tilde{x}_t} \log q(\tilde{x}_t|S_i) - \nabla_{\tilde{x}_t} \log q(\tilde{x}_t))\\
  \Rightarrow\; &\epsilon_\theta(\tilde{x}_t, t,\{S_i\}_{i=1}^L) = \epsilon_\theta(\tilde{x}_t, t, \emptyset) + \sum_{i=1}^L \Big(\epsilon_\theta(\tilde{x}_t, t,S_i) - \epsilon_\theta(\tilde{x}_t, t, \emptyset)\Big).
\end{align*}
By these equations, we can sample from $q(\tilde{x}_0|\{S_i\}_{i=1}^L)$ with classifier-free guidance using the perturbed noise:
\begin{align*}
    {\epsilon}^* &\coloneqq \epsilon_\theta(\tilde{x}_t, t, \emptyset) + \omega\Big(\epsilon_\theta(\tilde{x}_t, t,\{S_i\}_{i=1}^L) - \epsilon_\theta(\tilde{x}_t, t, \emptyset)\Big)\\
    &= \epsilon_\theta(\tilde{x}_t, t, \emptyset) + \omega\sum_{s\in \mathcal{D}_j}^L \Big(\epsilon_\theta(\tilde{x}_t, t,s) - \epsilon_\theta(\tilde{x}_t, t, \emptyset)\Big).
\end{align*}

We borrowed this derivation from \cite{liu2022compositional} to support the completeness of our work.
Although the integration of conditioning styles $\{S_i\}_{i=1}^L$ requires them to be conditionally independent given the generated sample $\tilde{x}_0$,  it has been observed that this condition does not strictly need to be strictly met in practice. 

\section{ Details of Diffusion Model}
  \label{append:diff}
We present a 1D UNet implementation that includes key components such as style embedding layer. During training, the model takes in a 1D time-series sample, an activity style vector, and a timestep to produce noise of the same dimension as the input. During sampling, the model uses noise, concatenated activity style vectors, and a timestep to generate a new time-series sample. Our diffusion model operates according to these specifications.

  \subsection{Architecture and Training Details}
\textbf{Architecture.} The model begins with an initialization convolution layer, followed by a series of downsampling blocks. Each downsampling block comprises two residual blocks and an attention layer, executed using a 1D convolutional layer with a kernel size of 3. The output of each downsampling block is saved in a list, which is later used in the upsampling process. After the downsampling blocks, the model has a middle block consisting of a residual block and an attention layer. The upsampling blocks are then applied in reverse order, with each block consisting of two residual blocks and an attention layer. The upsampling operation is performed using a transposed convolutional layer with a kernel size of 3. In addition to the convolutional layers and residual blocks, the model also includes a time embedding layer, which consists of a sinusoidal positional embedding and two linear layers with a channel size of 256. We borrow the code for the 1D UNet from \url{https://github.com/lucidrains/denoising-diffusion-pytorch}. Differently, we add a style embedding layer, which consists of a linear layer with a channel size of 100 and a linear layer with a channel size of 64. Both of these embedding layers are concatenated along the channel dimension, resulting in a tensor that is used as the condition for each residual block in the UNet.

\textbf{Training details.}
Our diffusion training settings primarily adhere to the guidelines outlined in \cite{ho2020denoising,ho2022classifier}. The batch size is 64, with a learning rate of $2 \times 10^{-4}$ using the Adam optimizer. The diffusion step is set to $T = 100$.  We choose the probability of dropping the conditioning information to be 0.5.
% The minimum noise level $\beta_1$ is $10^{-4}$ and maximum noise level $\beta_t$ is 0.2. We use exponential moving average (EMA) on model parameters with a decay factor of $0.9999$. 

\subsection{Sampling Procedure}
Once our diffusion is trained, we can use the style-fused sampling strategy to diversify the condition space for our task. During the selection of style sets for generating samples, we specify the ratio $\kappa$ of new samples to original samples, as well as the maximum number $o$ of styles that can be fused for each new sample and the number of fused styles is determined by a random variable that follows a specified distribution.

For example, if we have 1000 training samples and $\kappa = 1$ and $o = 5$, we will generate 1000 training samples.
For each new sample, we can constrain it to be fused with up to 5 styles, and the number of fused styles is determined by a random variable that follows a specified distribution. The distribution used to determine the number of fused styles can be customized based on the specific task and dataset.
For example, we can use a uniform distribution to ensure an equal probability of fusing any number of styles, or we can use a Poisson distribution to favor fewer fused styles. 
In our code, we have set a default probability distribution for the number of fused styles. For example, when $o = 5$ , we set the probability distribution for mixing from 1 to 5 styles to \{0.2, 0.2, 0.2, 0.2, 0.2\}, with a total sum of 1. This means that each new sample can be fused with up to 5 styles, and the number of fused styles is determined randomly based on this distribution.
However, this probability distribution can be flexibly adjusted to potentially achieve better results. By adjusting these hyperparameters, we can control the number of fused styles for each new sample and explore different combinations of styles. This could help us generate more diverse and representative samples for our task.

\subsection{Implementation of Generation in TSC models}
To leverage parallel computing for training time series classification (TSC) models, we simultaneously process a batch of training data containing $B$ samples and generate $\kappa \times B$ new samples. Within each batch, we select an appropriate number of style sets for each class in a class-balanced manner and aggregate all $\kappa \times B$ style sets as conditional inputs.
Next, we use our diffusion model to generate $\kappa \times B $ new samples. Once this batch generation process is complete, we utilize the generated samples to train the feature extractor, while the generated data with its class labels are stored.
This generation process is repeated for each batch of training data, requiring execution only once. The stored data will be directly used to train the feature extractor in subsequent epochs without the need for data regeneration.

\textbf{Budget.} Overall, our approach offers a cost-effective solution to the challenges associated with human activity data collection in HAR scenarios. For instance,  the cost of a three-axis accelerometer typically ranges from several tens of dollars, and collecting human activity data for 30,000 samples of various activities can take several weeks and cost approximately \$1,000 per participant. Additionally, manual annotation of the data in subsequent stages can lead to additional expenses. In contrast, our diffusion model only uses a GPU like RTX 3090 to create over 30,000 labeled activity samples in just one hour \footnote{Renting a single RTX 3090 GPU for one hour typically costs less than \$0.2} . These generated samples can provide significant performance gains for various baselines. Importantly, our method only requires a one-time expansion without the need for re-generation. Moreover, the generated samples may simulate new and unseen users, making the trained deep learning models more likely to be effectively deployed on new edge devices for real-world applications.

\section{Diversity learning strategy}
\renewcommand{\thefootnote}{\arabic{footnote}}
\label{append:stra}
 \subsection{Details of training TSC models}
After generating synthetic dataset $\tilde{D}^s$, the data space expands to ${D}^{s} = \{\tilde{D}^s \cup D^s\} = \{(\mathbf{X}_i, y^c_i)\}_{i=1}^{n^s + \tilde{n}^s}$ \footnote{Here, $y^c_i$ denotes the class label, which has the same meaning as $y_i$ in the paper.}. To enhance the diversity of learned features from ${D}^{s}$, we present a simple yet effective diversity learning strategy that is adapted from the representation learning method proposed in \cite{qin2023generalizable}.Our approach involves differentiating between the origin of each sample, whether it is "synthetic" or "original," and its corresponding class label. Since our augmented data is highly diverse, we adopt a simplified learning objective that removes complex computations, such as measuring the distance and similarity between synthetic and original data. Instead, we depend entirely on the cross-entropy loss, where each loss is determined by different classification criteria.
To this end, we employ a multi-objective method that consists of three sequential steps to train the classifier on  ${D}^{s}$. 

Specifically, there are three fundamental components of a TSC model: the feature extractor $G_f$, the projection layer $G_{\text{proj}^{\text{*}}}$, and the classifier $G_{y^{\text{*}}}$. The function $G_f  (\cdot)$ maps the inputs to their respective representations and is updated throughout all three steps. The function $G_{\text{proj}^{\text{*}}} (\cdot)$ is a fully connected layer that maps the representations to  vectors of length $Z$. The function $G_{y^{\text{*}}} (\cdot)$ is a classifier responsible for predicting the designed label. The superscripts (*) indicate that these components are utilized in different steps. The three steps are then described as follows:

%\noindent
\textbf{(i) Class-origin feature learning.}
To  learn more detailed representations, we label each sample based on its origin and class. The original data is labeled as $y^{\text{o}}_i = 1$ and the augmented data as  $y^{\text{o}}_i = 0$ . We then combine the origin and class labels to create new labels, represented as $y^{\text{co}}_i = (y_i^{\text{c}} + y^{\text{o}}_i \times C ) \in\mathbb{N}$. By using the classifier  $G_{y^{\text{co}}}: \mathbb{R} ^{Z} \rightarrow \mathbb{R} ^{2 \times C}$, we train the model to predict these new labels using cross-entropy, which can be expressed as:
\begin{equation}
  \mathcal{L}_\text{cls-ori}  = \frac{1}{n^s + \tilde{n}^s} \sum_{i=1}^{n^s + \tilde{n}^s}  \ell\left( G_{y^{\text{co}}}\left( G_{\text{proj}^{\text{co}}}\left( G_f(\mathbf{X}_i)\right)\right), y^{\text{co}}_i\right).
\end{equation}

%\noindent
\textbf{(ii) Origin-specific feature learning.}
To further enhance the distinction between synthetic and original data, we encourage the model to differentiate between origin labels using the loss function $L_{\text{ori-spe}}$, which is defined as follows:
\begin{equation}
  \mathcal{L}_\text{ori-spe}  = \frac{1}{n^s + \tilde{n}^s} \sum_{i=1}^{n^s + \tilde{n}^s}  \ell\left( G_{y^{\text{o}}}\left( G_{\text{proj}^{\text{o}}}\left( G_f(\mathbf{X}_i)\right)\right), y^{\text{o}}_i\right).
  \end{equation}
Here, $G_{y^{\text{o}}}: \mathbb{R} ^{Z} \rightarrow \mathbb{R} ^{2}$ serves as an origin classifier to distinguish whether the input features originate from a synthetic or an original sample. 

%\noindent 
\textbf{(iii) Class-specific feature learning.} The feature extractor finally undergoes a training process on ${D}^s$ to correctly predict the class labels. This step allows the model to separate clusters belonging to different classes. We employ a class classifier $G_{y^{\text{c}}} :\mathbb{R} ^{Z} \rightarrow \mathbb{R} ^{C}$ by minimizing the following loss:
\begin{equation}
  \mathcal{L}_\text{cls-spe}  = \frac{1}{n^s + \tilde{n}^s} \sum_{i=1}^{n^s + \tilde{n}^s}  \ell\left( G_{y^{\text{c}}}\left( G_{\text{proj}^{\text{c}}}\left( G_f(\mathbf{X}_i)\right)\right), y^{\text{c}}_i\right).
  \end{equation}

% \noindent
\textbf{Overall.} 

During the training process, as Algorithm \ref{alg:training} shows, the three steps are iteratively repeated in each epoch until termination. In the inference phase, as Algorithm \ref{alg:inference} shows, we only use the trained feature extractor $G_f$, projection layer $G_{\text{proj}^{\text{c}}}$ and the class classifier $G_{y^{\text{c}}}$ from step (iii). They are stacked together to classify the input time-series samples in the test dataset $D^t$.  

\subsection{Pseudo-code}

\begin{algorithm}[H]
  \caption{Training Algorithm for Diversity Learning Strategy}
  \label{alg:training}
  \begin{algorithmic}[1]
  \STATE \textbf{Initialize:} $G_f$ (feature extractor), $G_{\text{proj}^{\text{co}}}$ (projection layer), $G_{y^{\text{co}}}$ (classifier)
  \STATE \textbf{Input:} $D^s$.
  \STATE \textbf{Output:} $G_f$, $G_{\text{proj}^{\text{c}}}$, $G_{y^{\text{c}}}$
  \FOR{each epoch}
      \FOR{each batch $\mathbf{B}_i$ in training dataset $D^s$}
      \IF{epoch is 0}
      \STATE \textbf{{{\emph{Generate}} $B \times \kappa$ new samples $\tilde{\mathbf{B}}_i$.}}
      \STATE \textbf{{{\emph{Expand}} data space $D^s = D^s \cup \tilde{\mathbf{B}}_i$ and $\mathbf{B}_i = \mathbf{B}_i \cup \tilde{\mathbf{B}}_i$.}}
      
  \ENDIF
  
          \STATE \textbf{Predict} class-origin label: $\tilde{y}^{\text{co}}_i = G_{y^{\text{co}}}(G_{\text{proj}^{\text{co}}}(G_f(\mathbf{B}_i)))$.
          \STATE \textbf{Calculate} class-origin  loss: $\mathcal{L}_{\text{cls-ori}}$.
          \STATE \textbf{{{Update} $G_f$, $G_{\text{proj}^{\text{co}}}$, and $G_{y^{\text{co}}}$ using $\mathcal{L}_{\text{cls-ori}}$}}.
          \STATE \textbf{Predict} origin label: $\tilde{y}^{\text{o}}_i = G_{y^{\text{o}}}(G_{\text{proj}^{\text{o}}}(G_f(\mathbf{B}_i)))$.
          \STATE \textbf{Calculate} original-specific  loss: $\mathcal{L}_{\text{ori-spe}}$.
          \STATE \textbf{{{Update} $G_f$, $G_{\text{proj}^{\text{o}}}$, and $G_{y^{\text{o}}}$ using $\mathcal{L}_{\text{ori-spe}}$}}.
          \STATE \textbf{Predict} class label: $\tilde{y}^{\text{c}}_i = G_{y^{\text{c}}}(G_{\text{proj}^{\text{c}}}(G_f(\mathbf{B}_i)))$.
          \STATE \textbf{Calculate} class-specific loss: $\mathcal{L}_{\text{cls-spe}}$.
          \STATE \textbf{{{Update} $G_f$, $G_{\text{proj}^{\text{c}}}$, and $G_{y^{\text{c}}}$ using $\mathcal{L}_{\text{cls-spe}} $}}.
      \ENDFOR
  \ENDFOR
  \end{algorithmic}
\end{algorithm}

\begin{algorithm}[H]
  \caption{Inference}
  \label{alg:inference}
  \begin{algorithmic}[1]
  \STATE \textbf{Input:} Trained models $G_f$,  $G_{\text{proj}^{\text{c}}}$, $G_{y^{\text{c}}}$; Input data $D^t$.
  \STATE \textbf{Output:} Predicted labels $\hat{Y}$.
  \FOR{each input sample $\mathbf{x}$ in $D^t$}
      \STATE Predict class label:  $\tilde{y}^{\text{c}}_i = G_{y^{\text{c}}}(G_{\text{proj}^{\text{c}}}(G_f(\mathbf{x})))$.
  \ENDFOR
  \end{algorithmic}
\end{algorithm}

\section{Details of Experimental Setup }
  \label{append:exp}
\subsection{Datasets}
To ensure fairness and reproducibility, we conduct our experiments by the same experimental setup as that provided in \cite{qin2023generalizable} to compare the performance of our method on HAR tasks. Our datasets for this study include DSADS \footnote{\url{https://archive.ics.uci.edu/dataset/256/daily+and+sports+activities}}, PAMAP2 \footnote{\url{https://archive.ics.uci.edu/dataset/231/pamap2+physical+activity+monitoring}}, and USC-HAD \footnote{\url{https://sipi.usc.edu/had/}}. We follow the same processing steps involving the domain split and randomly choose remaning data used for each dataset, as detailed in the official code \footnote{\url{https://github.com/microsoft/robustlearn/tree/main/ddlearn}}. The corresponding processing information is as follows:

\textbf{Dataset information and pre-processing.}
The information details of the three  HAR datasets are presented in Table~\ref{tb-dataset}.
To segment the data, we employ a sliding window approach. For DSADS, the window duration is set to 5 seconds as per~\cite{barshan2014recognizing}, while for PAMAP2, it is set to 5.12 seconds~\cite{reiss2012introducing}. Similarly, for USC-HAD, a 5-second window is utilized with a 50\% overlap between consecutive windows. Given the sampling rates of each dataset (25Hz for DSADS, 100Hz for PAMAP2, and 100Hz for USC-HAD), the window lengths are calculated to be 125 readings, 512 readings, and 500 readings, respectively. We normalize the data using normalization and reshape the 1D time series sample into a 2D format with a height of 1. Each batch is structured as $(b, c, h, w)$, where $b$ represents the mini-batch size, $c$ denotes the number of channels corresponding to the total axes of sensors, $h$ signifies the height, and $w$ denotes the window length.

\textbf{Domain split.}
We implement leave-one-out-validation by dividing subjects into multiple groups. In this approach, we designate one group of subjects' data as the target domain, while the remaining subjects' data serve as the source domain. Each group can be considered as an individual task. For DSADS and PAMAP2, we divide the 8 subjects into 4 groups. As for USC-HAD, we divide the 14 subjects into 5 groups, with groups 0-3 comprising 3 subjects each, and the last group consisting of 2 subjects. Subsequently, we partition the data within each group into training, validation, and test sets, maintaining a ratio of 6:2:2.
To assess the impact of training data size on model performance, we conduct experiments where we randomly sample 20\% to 100\% of the training data with increments of 20\%. This allows us to simulate the small-scale setting. During testing, we evaluate the trained model on the test set of the target domain. 

\begin{table*}[htbp]
  \caption{Summary of information and feature network settings for three HAR datasets}
  \label{tb-dataset}
  \centering
  \resizebox{0.8\textwidth}{!}{
  \begin{tabular}{cccccccc}
  \toprule
  Dataset & Task & Subjects & Activities ($C$) & Sample Shape & Kernel Size & Network (feature and projection) & Output Channel \\ 
  \midrule
  DSADS   & 4 & 8         & 19 & (45, 1, 125) & 9 & \makecell{2 Conv1D Layers  \\ 1 FC Layer } & (16, 32, 64) \\ 
  PAMAP2  & 4 & 9         & 8 & (27, 1, 512) & 9 & \makecell{2 Conv1D Layers\\ 1 FC Layer} & (16, 32, 64) \\ 
  USC-HAD & 5 & 14        & 12 & (6, 1, 500) & 6 & \makecell{3 Conv1D Layers  \\ 1 FC Layer } & (16, 32, 64, 128) \\ 
  \bottomrule
  \end{tabular}}
\end{table*}
\subsection{Baselines}
We compared our proposed model with several closely related DG baselines that are also suitable for time series input:
\begin{tight_itemize}

  \item \texttt{Mixup} \cite{xu2020adversarial}: Through appwalking domain mixup at both pixel and feature levels, it is a data
  augmentation-based DG method.
  \item \texttt{RSC} \cite{huang2020self}: A training method enhances out-of-domain generalization by discarding dominant features in the training data.
  \item \texttt{SimCLR} \cite{chen2020simple}: It leverages data augmentation to generate positive samples and a learnable transformation for contrastive learning.
  \item \texttt{Fish} \cite{shi2021gradient}: By maximizing the gradient inner product between domains, it learns domain-invariant features in DG issues.
  \item \texttt{DANN} \cite{ganin2016domain}: It uses adversarial training to learn domain-invariant features from data with accessible target labels.
  
  \item \texttt{mDSDI} \cite{bui2021exploiting}: It optimizes domain-specific features via meta-learning from source domains, where the domain labels are known.
  \item \texttt{TS-TCC} \cite{eldele2021time}: As a recent self-supervised method, it leverages a small set of labeled time-series data for model generalization.
  
  \item \texttt{DDLearn} \cite{qin2023generalizable}: A latest DG method that leverages standard data augmentation baselines, designed for low-resource scenarios.
\end{tight_itemize}
In our experiments, we use the same data augmentation techniques \cite{um2017data} as those used in \cite{qin2023generalizable} to ensure a fair comparison. These techniques include rotation, permutation, time-warping, scaling, magnitude warping, jittering, and random sampling. For example, jittering involves applying different types of noise to the samples, which increases the diversity of data magnitude. Scaling, on the other hand, rescales the samples to different magnitudes.

\subsection{Architecture}
All DG baselines use the same network architecture for feature extraction, except for TS-TCC \cite{eldele2021time}. Specifically, we adopt an architecture consisting of a feature extractor, a projection layer, and a classifier, as described in \cite{qin2023generalizable}. The feature extractor is composed of two or three Conv1D layers, depending on the dataset being used. For DSADS and PAMAP2, we use two Conv1D layers with a kernel size of 9, while for USC-HAD, we use three Conv1D layers with a kernel size of 6. Each Conv1D layer is followed by a ReLU activation function and a maxpool1d operation. The output is then connected to a projection layer, which is a fully-connected layer with output feature dimensions of 64 for DSADS and PAMAP2, and 128 for USC-HAD. Finally, we employ a fully-connected layer as the classifier, which takes the extracted features as input and outputs $C$-dimensional logits. By applying a softmax operation, we obtain prediction probabilities for each class, which sum up to 1.
In our diversity learning strategy, we use different projection layers and classifiers at different training steps, while keeping the overall architecture of the model the same. Specifically, we use different projection layers with the same architecture for different steps, and classifiers with the same input channel but output $(2 \times C)$-dimensional, $2$-dimensional, and $C$-dimensional logits for specific classification goals. During the inference phase, we utilize only a trained feature extractor, a projection layer, and a classifier.
Since TS-TCC is designed for contrastive learning, we retain its architecture and make slight modifications to adapt it to the input data channels and lengths for use in HAR tasks.

\subsection{Training Details}
All methods in our experiments use PyTorch.  We utilize Adam optimization with a scheduler. The batch size is fixed at 64. The experiments are conducted on one GeForce RTX 3090 Ti GPU. 
To ensure its performance, we fine-tune the training configurations for each baseline.  Furthermore, the results for \texttt{Mixup} \cite{xu2020adversarial}, \texttt{RSC} \cite{huang2020self}, \texttt{SimCLR} \cite{chen2020simple}, \texttt{Fish} \cite{shi2021gradient}, and \texttt{DDLearn} \cite{qin2023generalizable}  reported in Tables \ref{tab-acc-all} and \ref{tab-rate} are obtained from the paper \cite{qin2023generalizable}.








\end{document}